\documentclass[10.5pt,onecolumn]{article}

\usepackage[a4paper,total={6in, 8in}]{geometry}
\usepackage{algorithm}
\usepackage{algorithmic}
\usepackage{geometry}

\usepackage{mathrsfs,amsmath,graphicx,epsfig,epstopdf,float,algorithm,algorithmic,,amsfonts,amssymb,url,amsfonts,amssymb,color,mathrsfs,amssymb,booktabs,multirow,bm}
\usepackage{enumerate} % cite,
\usepackage{amsthm}
\usepackage{amssymb}

\theoremstyle{plain}
\newtheorem{As}{Assumption}

\theoremstyle{plain}
\newtheorem{Remk}{Remark}

\theoremstyle{plain}

\theoremstyle{plain}
\newtheorem{Lem}{Lemma}

\theoremstyle{plain}
\newtheorem{Thm}{Theorem}
\theoremstyle{plain}
\newtheorem{Prop}{Proposition}

\def \bbtheta {{\boldsymbol \theta}}
\def \bbphi {{\boldsymbol \phi}}
\def \BPhi {{\boldsymbol \Phi}}
\def \BTheta {{\boldsymbol \Theta}}
\def \DElta {{\boldsymbol \Delta}}

\allowdisplaybreaks

\begin{document}

\title{\Large
\textbf{Finite-Sample Analysis of Decentralized Temporal-Difference Learning  with Linear Function Approximation}
}

\author{Jun Sun, Gang Wang, Georgios B. Giannakis, Qinmin Yang, and Zaiyue Yang\thanks{The work by J. Sun and Z. Yang was supported in part by NSFC Grants 61873118, 61673347,
%the Shenzhen Committee on Science and Innovations under Grant GJHZ20180411143603361, 
and the Dept. of Science and Technology of Guangdong Province under Grant 2018A050506003.
 The work by J. Sun was also supported by the China Scholarship Council. The work by G. Wang and G. B. Giannakis was supported in part by NSF grants 1711471, and 1901134. The work of Q. Yang was supported in part by NSFC grants 61673347, U1609214, 61751205, 
%Key R\&D Program of Guangdong Province (No. 2018B010107002), 
and the Key R\&D Program of Zhejiang Province under Grant 2019C01050.
J. Sun and Q. Yang are with the College of Control Science and Engineering, and the State Key Laboratory of Industrial Control Technology, Zhejiang University, Hangzhou, China. G. Wang and G. B. Giannakis are with the Digital Technology Center and the Department of Electrical and Computer Engineering, University of Minnesota, Minneapolis, MN 55455, USA. Z. Yang is with the Department of Mechanical and Energy Engineering, Southern University of Science and Technology, Shenzhen, China. E-mail:  
junsun16sj@gmail.com; 
 gangwang@umn.edu;
 georgios@umn.edu;
 qmyang@zju.edu.cn;
 yangzy3@sustech.edu.cn;
}
%\thanks {Corresponding author: Zaiyue Yang, Email: yangzy3@sustech.edu.cn. }
}

\maketitle

\begin{abstract}
Motivated by the emerging use of multi-agent reinforcement learning (MARL) in engineering applications such as networked robotics, swarming drones, and sensor networks, we investigate the 
policy evaluation problem
 in a fully decentralized setting, using temporal-difference (TD) learning with linear function approximation to handle large state spaces in practice. The goal of a group of agents is to collaboratively learn the value function of a given policy from locally private rewards observed in a shared environment, through exchanging local estimates with neighbors. Despite their simplicity and widespread use, our theoretical understanding of such decentralized TD learning algorithms remains limited. 
 Existing results were obtained based on i.i.d. data samples, 
 or by imposing an `additional' projection step to control the `gradient' 
bias incurred by the Markovian observations. 
 In this paper, we provide a finite-sample analysis of the fully decentralized TD(0) learning under both i.i.d. as well as Markovian samples, and prove that all local estimates converge linearly to a small neighborhood of the optimum.
 The resultant error bounds are the first of its type---in the sense that they hold under the most practical assumptions
 ---which is made possible by means of a novel multi-step Lyapunov analysis.

\end{abstract}

%\begin{keywords}
%	Escaping local minima, 
%\end{keywords}

\section{INTRODUCTION}
Reinforcement learning (RL) is concerned with how artificial agents ought to take actions in an unknown environment so as to maximize some notion of a cumulative reward. Thanks to its generality, RL has been widely studied in many areas, such as control theory, game theory, operations research, multi-agent systems, machine learning, artificial intelligence, and statistics \cite{book2008rl}.
 In recent years, combining with deep learning, RL has demonstrated its great potential in addressing challenging practical control and optimization problems \cite{2015drl,
 	shalev2016safe,tsg2020yang,tccn2019}.
% 	liu2017reinforcement % 	franccois2018introduction, 
Among all possible algorithms, the temporal difference (TD) learning has arguably become one of the most popular RL algorithms so far, 
% lies at the core of RL, 
 which is further dominated by the celebrated TD(0) algorithm \cite{1988td}. 
%With the popularity of RL is increasing research interests, among which temporal difference (TD) learning attracts much attention. 
TD learning provides an iterative process to update an estimate of the so-termed value function $v^{\mu}(s)$ with respect to a given policy $\mu$ based on temporally successive samples. 
%In this paper, we will particularly focus on value function estimation, or, policy evaluation, that is, evaluating the value function of each state with the policy fixed. 
Dealing with a finite state space, 
the classical version of the TD(0) algorithm adopts a tabular representation for $v^{\mu}(s)$, which stores
entry-wise value estimates on a per state basis.
 
Although it is conceptually simple as well as easy-to-implement, the tabular TD(0) learning algorithm can become intractable when the number of states grows large or even infinite, which emerges in many contemporary control and artificial intelligence problems of practical interest. This is also known as the ``curse of dimensionality'' \cite{book1996ndp}.  
%meaning value function estimation with TD learning becomes intractable when the state space grows large.
The common practice to bypass this hurdle, is to approximate the exact tabular value function with a class of function approximators, including for example, linear functions 
%as $v_{\pi}(s)\approx \bm{\theta}^\top\phi(s)$ 
or nonlinear ones using even deep neural networks \cite{book2008rl}. 
%In this context, the parametric approximation of the value function is essential for large-scale applications.
% Generally, the approximation can be either nonlinear, say, using neural network or linear.

%Although nonlinear approximators can be much more powerful, linear function approximation allows for an efficient implementation of TD(0) learning even on large state spaces, which has been demonstrated to perform well in a variety of applications \cite{powell2007approximate}, \cite{book2008rl}.  
%As a matter of fact, recent theoretical RL efforts have mostly focused on linear function approximation; see e.g., \cite{1988td,1994tdc,1995tdd,2015tdf,2017error,2017gtd,2015tdf,2018go,bhandari2018finite,nips2019hu}. Along this line, we will also focus on TD(0) learning with linear function approximation, and particularly aim at developing non-asymptotic performance guarantees for TD(0) learning in a completely decentralized setting. 
%Nonetheless, we consider this work as a preliminary milestone in route to achieving theoretical guarantees for non-linear RL architec- tures.
%  Our interests fall in the linear approximation in hope to simplify the theoretical analysis and work as a stepping stone for the analysis of nonlinear counterpart.

Albeit nonlinear function approximators using e.g., deep neural networks \cite{2015drl,wang2019relus}, can be more powerful, linear approximation allows for an efficient implementation of TD(0) even on large 
or infinite
 state spaces, which has been demonstrated to perform well in a variety of 
%real-world
 applications \cite{powell2007approximate}, \cite{book2008rl}.  
Specifically, TD learning with linear function approximation parameterizes the value function with a linear combination of a set of preselected basis functions (a.k.a., feature vectors) induced by the states, and estimates the coefficients in the spirit of vanilla TD learning. 
Indeed, recent theoretical RL efforts have mostly centered around linear function approximation; see e.g., \cite{1994tdc,1995tdd,
%	2017error,
%	2017gtd,
%	2015tdf,
	bhandari2018finite,nips2019hu,gupta2019finite,xu2020reanalysis}. 
%Along this line, we will also focus on TD(0) algorithms with linear function approximation in this paper, and particularly aim at developing non-asymptotic performance guarantees for TD(0) learning  in a completely decentralized setting. 

%Although the dimension of the parameter can be moderate compared with large state space in practical applications, the performance of linear approximation has been recognized as promising \cite{powell2007approximate}. Nevertheless, the theoretical analysis of TD learning has long remained subtle since it was proposed. 

Early theoretical convergence results of TD learning were mostly asymptotic \cite{1988td,1994tdc,1995tdd,pananjady2019value}; that is, results that hold only asymptotically when the number of updates (data samples) tends to infinity. 
%This is because TD learning does not minimize a fixed explicit objective as conventional optimization algorithms do. 
By exploring the asymptotic behavior, TD(0) learning with linear function approximation can be viewed as a discretized version of an ordinary differential equation (ODE) \cite{580874}, or a linear dynamical system \cite{book2009sa}, so TD(0) updates can be seen as tracking the trajectory of the ODE provided the learning rate is infinitely small \cite{580874}. Indeed, this dynamical systems perspective has been widely used to study the asymptotic convergence of general stochastic approximation algorithms \cite{book2009sa}.
%, thus the ultimate goal corresponds to solving the optimization problem associated with the ODE. 
%Building on this idea, a general framework for the convergence of stochastic approximation is built, which can be applied to the convergence of TD learning \cite{book2009sa}. 
 Motivated by the need for dealing with massive data in modern signal processing, control, and artificial intelligence tasks (e.g., \cite{tsp2019chi,2015drl}), recent interests have centered around developing non-asymptotic performance guarantees that hold with even finite data samples and help us understand the efficiency of the algorithm or agent in using data. 
 
%Although the ODE approach provides only asymptotic convergence guarantees, 
%but cannot derive the convergence rate for finite time setting. 

Non-asymptotic analysis of RL algorithms, and TD learning in particular, is generally more challenging than their asymptotic counterpart, due mainly to two reasons that: i) TD updates do not correspond to minimizing any static objective function as standard optimization algorithms do; and, ii) data samples garnered along the trajectory of a single Markov chain are correlated across time, resulting in considerably large (possibly uncontrollable) instantaneous `gradient' bias in the updates. Addressing these challenges, a novel suite of tools has lately been put forward.    
A convex-concave saddle-point formulation was introduced by \cite{liu2015finite} %touati2018convergent
 to facilitate finite-time analysis.
 of a TD variant, termed gradient (G) TD with linear function approximation. 
Adopting the dynamical system viewpoint, the 
%(running-average of)
iterates of TD(0) updates after a projection step were shown converging to the equilibrium point of the associated ODE at a sublinear rate in  \cite{dalal2018td}.
%, where a high probability bound is also provided. 
With additional transformation and/or projection steps,
% to the parameter estimates, 
finite-time error bounds of a two-timescale TD learning algorithm developed by \cite{sutton2009convergent} were established in
 \cite{gupta2019finite,2019xu}. 
 The authors in \cite{bhandari2018finite} unified finite-time results of TD(0) with linear function approximation, under both identically, and independently distributed (i.i.d.) noise, as well, as Markovian noise. 
 % However, these aforementioned non-asymptotic results were all developed based on modified variants of TD(0).

%In summary, recent studies have attained a variety of results on the finite analysis of TD learning; however, 

In summary, these aforementioned works in this direction either assume i.i.d. data samples \cite{dalal2018td}, or have to incorporate a projection step
  \cite{bhandari2018finite}. As pointed out in
\cite{dalal2018td} however, although widely adopted, i.i.d. samples are difficult to acquire in practice.
On the other hand, the projection step is imposed only
% to bound the gradient bias
 for analysis purposes, which requires prior knowledge to select the size of a feasibility set.  
% prevent us from obtaining a better solution from the  whole range of feature space.
Moreover, most existing theoretical RL studies have considered the centralized setting, except for e.g., \cite{wai2018multi,doan2019finite}
concerning theoretical aspects of decentralized RL under the i.i.d. assumption and/or with the projection step; while early efforts on multi-agent RL focused on empirical performances \cite{foerster2016learning}. In a fully decentralized setting, multi-agents share a common environment but observe private rewards. With the goal of jointly maximizing the total accumulative reward, each agent can communicate with its neighbors, and updates the parameter locally. Such decentralized schemes appear naturally in numerous applications, including, for instance, robotics \cite{yan2013survey}, mobile sensor networks \cite{krishnamurthy2008decentralized}, and drone control \cite{yanmaz2017communication}. 
%Prior works on centralized setting still cannot get rid of the I.I.D assumption or the projection \cite{zhang2018fully,doan2019finite}.

As a complementary to existing theoretical RL efforts, this paper offers a novel finite-sample analysis for a fully decentralized TD(0) algorithm with linear function approximation.
 For completeness of our analytical results, we investigate both the i.i.d. case as, well as, the practical yet challenging Markovian setting, where data samples are gathered along the trajectory of a single Markov chain. 
% Our results are the first of their kind in both i.i.d and Markovian settings. 
%with the goal of extending the I.I.D observation to more practical Markovian observation and removing the projection.
 With communications of local parameter estimates between neighbors, we first establish consensus among all agents. To render the finite-time analysis under the Markovian noise possible, we invoke a novel multi-step Lyapunov approach \cite{wang2019multistep}, which successfully eliminates the need for a projection step as required by \cite{doan2019finite}. Our theoretical results show that a fully decentralized implementation of the original TD(0) learning, converges linearly to a neighborhood of the optimum under both i.i.d. and Markovian observations. Furthermore, the size of this neighborhood can be made arbitrarily small by choosing a small enough stepsize. 
 In a nutshell, the main contributions of this paper are summarized as follows.
\begin{enumerate}
\item[c1)] We investigate the fully decentralized TD(0) learning with linear function approximation, and establish the multi-agent consensus, as well as their asymptotic convergence; and,
\item[c2)] We provide finite-time error bounds for all agents' local parameter estimates in a fully decentralized TD(0) setting, under both i.i.d. and Markovian observations, through a multi-step Lyapunov analysis.  
% all local parameters will reach consensus and converge to at a linear rate to the neighborhood of optimal solution. The theoretical analysis is built on  Markovian observation and does not depend on bounded projection of parameter.
\end{enumerate}

%\textbf{Notation}. Calligraphic letters denote sets, and bold lower- (upper-)case letters denote column vectors (matrices), e.g., $\bm{x}$ ($\bm{X}$); $ x_i$ represents the $i$-th element of vector $\bm{x}$, and correspondingly, $ X_{ij}$ the $ij$-th entry of matrix $\bm{X}$; $\|\bm{x}\|$ ($\|\bm{X}\|_{F}$) stands for the $\ell_2$-norm (respectively, Frobenious norm) of $\bm{x}$ ($\bm{X}$);
%%; while $\left\lfloor a\right\rfloor$ denotes downward rounding of $ a $; and 
% $|\cdot|$ denotes the absolute value of a scalar or the cardinality of a set clear from the context; and $ \bm{I}$ denotes identity matrices of suitable dimensions, while symbols $ \bm{0}$ $ (\bm{1})$ represents an all-zero (all-one) vector of proper size.  

\section{DECENTRALIZED REINFORCEMENT LEARNING}
A discounted Markov decision process (MDP) is a discrete-time stochastic control process, which can be characterized by a $5$-tuple $\small (\mathcal{S}, \mathcal{A}, P^a, R^a, \gamma)$. Here, $\small \mathcal{S}$ is a finite set of environment and agent states, $\small \mathcal{A}$ is a finite set of actions of the agent, $ \small P^a(s,s')={\rm Pr}(s'|s,a)$ is the probability of transition from state $\small s\in \mathcal{S}$ to state $\small s'$ upon taking action $\small a\in\mathcal{A}$, $\small R^a(s,s'): \mathcal{S}\times\mathcal{S}\rightarrow \mathbb{R}$ represents the immediate reward received after transitioning from state $s$ to state $s'$ with action $a$, 
%which is assumed uniformly bounded by $\small r_{\max}>0$, 
and $\small \gamma$ is the discounting factor.

The core problem of MDPs is to find a policy for the agent, namely a mapping $\mu:\mathcal{S}\times\mathcal{A}\to [0,1]$ that specifies the probability of choosing action $a\in\mathcal{A}$ when in state $s$. 
%Here, we focus on a given policy $\mu$.
 Once an MDP is combined with a policy, this fixes the action for each state and their combination determines the stochastic dynamics of a Markov chain \cite{maei2009td}.
%  the resultant combination is a Markov chain.
  Indeed, this is because the action $a$ chosen in state $s$ is completely determined by $\mu(s,a)$, 
then ${\rm Pr}(s'|s,a)$
 reduces to $P^{\mu}(s,s')=\sum_{a\in\mathcal{A}} \mu(s,a) P^a(s'|s)$, a Markov transition matrix $\bm{P}^{\mu}$. Likewise, immediate reward $R^a(s,s')$ also simplifies to the expected reward $R^{\mu}(s,s')=
% \sum_{s'\in\mathcal{S}}
 \sum_{a\in\mathcal{A}} \mu(s,a) P^a(s'|s) R^a(s'|s)$.

The quality of policy $\mu$ is evaluated in terms of the expected sum of discounted rewards over all states in a finite-sample path while following policy $\mu$ to take actions, which is also known as the value function $ v^{\mu}: \mathcal{S}\rightarrow\mathbb{R}$.
%In this paper, we focus on the fixed policy $\mu$, following which the value function $\small v^{\mu}(s): \small \mathcal{S}\rightarrow\mathbb{R}$ summing the expected discounted rewards starting from state $\small s$ can be denoted by $\small v(s)$ (
In this paper, we focus on evaluating a given policy $\mu$, so we will neglect for notational brevity the dependence on $ \mu$ hereafter.
Formally, $\small v(s)$ is defined as follows
\begin{equation}\label{eq:value}
v(s)=\mathbb{E}\bigg[\sum_{k=0}^{\infty}\gamma^k R(s(k),s(k+1))\Big|s(0)=s\bigg],~~  \forall s\in\mathcal{S}
\end{equation}
where the expectation is taken over all transitions from $\small k=0$ to $\small k=+\infty$.

Assuming a canonical ordering on the elements of $\mathcal{S}$, say a renumbering $\{1,2,\ldots,|\mathcal{S}|\}$,
we can treat $ v$ as a $|\mathcal{S}|$-dimensional vector $\bm{v}:=[v(1)~\, v(2)\, \cdots\, v(|\mathcal{S}|) ]^\top\in\mathbb{R}^{|\mathcal{S}|}$.
%, and $\bm{P}$ as a $|\mathcal{S}|\times |\mathcal{S}|$ Markov transition matrix. 
It is well known that the value function $v(s)$ in \eqref{eq:value} satisfies the so-called Bellman equation \cite{book1996ndp}
% equation \eqref{eq:value} can be written in an alternative way, that is, written as Bellman equation:
\begin{equation}\label{eq:bellman}
v(s)=\sum_{s'\in\mathcal{S}}P_{ss'}\big[R(s,s')+\gamma v(s')\big], ~~\forall s\in\mathcal{S}.
\end{equation}
If the transition probabilities $\{P_{ss'} \}$ and the expected rewards $\{ R(s,s')\}$ were known, finding $v\in\mathbb{R}^{|\mathcal{S}|}$ is tantamount to solving a system of linear equations described by \eqref{eq:bellman}.   
It is obvious that when the number of states $ |\mathcal{S}|$ is large or even infinite, exact computation of $v$ can become intractable, which is also known as the ``curse of dimensionality'' \cite{book1996ndp}. This thus motivates well a low-dimensional (linear) function approximation of $\small v(s)$, parameterized by an unknown vector $\bm{\theta}\in\mathbb{R}^p$ as follows
\begin{equation}
v(s)\approx \tilde{v}(s,\bbtheta)=\bbphi^\top(s)\bbtheta,~~\forall s\in\mathcal{S}\label{eq:linapp0}
\end{equation}
where we oftentimes have the number of unknown parameters $p\ll |\mathcal{S}|$; and $\small \bbphi(s)\in\mathbb{R}^{p}$ is a preselected feature or basis vector characterizing state $\small s\in\mathcal{S}$. 

%can be used to facilitate the generalization of observed transitions to unvisited or rarely visited states

For future reference, let vector
$ \tilde{\bm{v}}(\bbtheta):=[\tilde{v}(1,\bbtheta)\,~ \tilde{v}(2,\bbtheta) \, $ $ \cdots\,  \tilde{v}(|\mathcal{S}|,\bbtheta) ]^\top$ collect the value function approximations at all states,  and define the feature matrix
\begin{equation*}
\BPhi:=\begin{bmatrix} \bbphi^\top(1)  \\ \bbphi^\top(2)  \\ \vdots  \\ \bbphi^\top(|\mathcal{S}|) \end{bmatrix}\in\mathbb{R}^{|\mathcal{S}|\times p}
\end{equation*}
 then it follows that
\begin{equation}\label{eq:linapp1}
 \tilde{\bm{v}}(\bbtheta)=\BPhi \bbtheta.
\end{equation}

Regarding the basis vectors $\{\bm{\phi}(s) \}$ (or equivalently, the feature matrix $\BPhi$), we make the next two standard assumptions \cite{580874}: i) $\|\bbphi(s)\|\le 1$, $\forall s\in\mathcal{S}$, that is, all feature vectors are normalized; and, ii) $\BPhi$ is of full column rank, namely, all feature vectors are linearly independent.

With the above linear approximation, the task of seeking $\bm{v}$ boils down to find the parameter vector $\bbtheta^*$ that minimizes the gap between the true value function $\bm{v}$ and the approximated one
 $\tilde{\bm{v}}(\bbtheta)$. Among many possibilities in addressing this task, the original temporal difference learning  algorithm, also known as TD(0), is arguably the most popular solution \cite{1988td}.
The goal of this paper is to develop decentralized TD(0) learning algorithms and further investigate their finite-time performance guarantees in estimating $\bbtheta^*$. 
To pave the way for decentralized TD(0) learning, let us start off by introducing standard centralized version below. 

\subsection{Centralized Temporal Difference Learning}
%Suppose that at time instant $k$, the environment transits from state $s(k)$ to state $s(k+1)$, and generates immediate reward $r(k)=R(s(k),s(k+1))$.
The classical TD(0) algorithm with function approximation \cite{1988td} starts with some initial guess $\bm{\theta}(0)\in\mathbb{R}^p$. Upon observing the $k^{\rm th}$ transition from state $s(k)$ to state $s(k+1)$ with reward $r(k)=R(s(k,s(k+1)))$, it first computes the so-called temporal-difference error, given by
\begin{equation}
d(k)=r(k)+\gamma \tilde{v}(s(k+1),\bbtheta(k))-\tilde{v}(s(k),\bbtheta(k))
\end{equation}
which is subsequently used to update the parameter vector $\bm{\theta}_k$ as follows
\begin{equation}\label{eq:tdupdate}
\bbtheta(k+1)=\bbtheta(k)+\alpha d(k) \nabla \tilde{v}(s(k), \bbtheta(k)).
\end{equation}
Here, $\alpha>0$ is a preselected constant stepsize, and the symbol $\nabla \tilde{v}(s(k),\bbtheta(k))=\bbphi(s(k))$ denotes the gradient of $\tilde{v}(s(k), \bbtheta)$ with respect to $\bbtheta$ evaluated at the current estimate $\bbtheta(k)$. 
For ease of exposition, we define the `gradient' estimate $\bm{g}(k)$ as follows
\begin{align}\label{eq:gradient}
\bm{g}(\bbtheta(k),\xi_k)&:=d(k) \nabla \tilde{v}(s(k), \bbtheta(k))\nonumber\\
&\,=\bbphi(s(k))\! \left[\gamma\bbphi^\top(s(k+1))-\bbphi^\top(s(k))\right]\bbtheta(k)+r(k)\bbphi(s(k)).
\end{align}
where $\xi_k$ captures all the randomness corresponding to the $k$-th transition $(s(k),s(k+1),\{r_m(k)\}_{m\in\mathcal{M}})$.
Thus, the TD(0) update \eqref{eq:tdupdate} can be rewritten as 
\begin{equation}\label{eq:tdde}
\bbtheta(k+1)=\bbtheta(k)+\alpha \bm{g}(\bbtheta(k),\xi_k).
\end{equation}

Albeit viewing $\bm{g}(\bm{\theta}(k),\xi_k)$ as some negative `gradient' estimate, the TD(0) update in \eqref{eq:tdde} based on online rewards 
%that calculates gradient and carries out online update based on current observation
resembles that of the stochastic gradient descent (SGD). It is well known, however, that even the TD(0) learning update does not correspond to minimizing any fixed objective function \cite{book2008rl}. Indeed, this renders convergence analysis of TD algorithms rather challenging, letting alone the non-asymptotic (i.e., finite-time) analysis.
    To address this challenge, 
 TD learning algorithms have been investigated in light of the stability of a dynamical system described by an ordinary differential equation (ODE) \cite{book2009sa,580874,wang2019multistep}.
% n ordinary differential equation (ODE) system.
%    tools from dynamical systems have been borrowed to study convergence of TD(0) learning algorithms.

Before introducing the ODE system for \eqref{eq:tdde}, let us first simplify the expression of $\bm{g}(\bbtheta(k))$.  Upon defining 
\begin{equation}\label{eq:hmatrix}
\bm{H}(\xi_k):=\bbphi(s(k))\big[\gamma\bbphi^\top(s(k+1))-\bbphi^\top(s(k))\big]
\end{equation}
and 
\begin{equation}
\bm{b}(\xi_k):=r(k)\bbphi(s(k))
\end{equation}
the gradient estimate $\bm{g}(\bbtheta(k))$ can be re-expressed as follows
\begin{equation}
\bm{g}(\bbtheta(k),\xi_k)=\bm{H}(\xi_k)\bbtheta(k)+\bm{b}(\xi_k). 
\end{equation}

Assuming that the Markov chain is finite,  irreducible, and aperiodic, there exists a unique stationary distribution $\bm{\pi}\in\mathbb{R}^{1\times |\mathcal{S}|}$ \cite{mcbook}, adhering to $\bm{\pi} \bm{P}=\bm{\pi}$. Moreover, let 
 $\bm{D}$ be a diagonal matrix holding entries of $\bm{\pi}$ on its main diagonal.
We also introduce $r'(s):=\sum_{s'\in\mathcal{S}}P(s,s')R(s,s')$ for all $s\in\mathcal{S}$ and collect them into vector $\bm{r}'=\big[r'(1)~r'(2)~\cdots~ r'(|\mathcal{S}|)\big]^\top$. 

It is not difficult to verify that after the Markov chain reaches the stationary distribution, then the following limits hold true
\begin{align}
\bar{\bm{H}}&:=\lim_{k\rightarrow \infty} \mathbb{E} [\bm{H}(\xi_k)]=\BPhi \bm{D}(\gamma \bm{P} \BPhi^\top-\BPhi^\top)\label{eq:hbar}\\
\bar{\bm{b}}&:=\lim_{k\rightarrow \infty} \mathbb{E} [\bm{b}(\xi_k)]=\BPhi \bm{D} \bm{r}'
\end{align}
yielding 
  \begin{equation} 
 \bar{\bm{g}}(\bbtheta):=\bar{\bm{H}}\bbtheta+\bar{\bm{b}}.
 \end{equation}

It has been shown that, under mild conditions on the stepsize $\alpha$, the TD(0) update \eqref{eq:tdupdate} or \eqref{eq:tdde} can be understood as tracking the following ODE  \cite{580874}
 \begin{equation}\label{eq:ode}
 \dot{\bbtheta}=\bar{\bm{g}}(\bbtheta).
 \end{equation}
%
%which indicates that as the time instant is large enough, meaning the MDP has arrived at its stationary distribution, TD(0) learning \eqref{eq:tdde} can be regraded as tracking the ODE \cite{580874}.
%Then, with $\bar{\bm{H}}$, $\bar{\bm{b}}$ and $\bar{\bm{g}}(\bbtheta)$ defined as 
%\begin{equation}
%\BPhi \bm{D}(\gamma P \BPhi-\BPhi), 
%\end{equation}
%and
%\begin{equation}
%\bar{\bm{b}}:=\BPhi \bm{D} \bm{r}',
%\end{equation}
% and
% \begin{equation} 
% \bar{\bm{g}}(\bbtheta):=\bar{\bm{H}}\bbtheta+\bar{\bm{b}},
% \end{equation}
% the ODE system is constructed as 
%\begin{equation}\label{eq:ode}
%\dot{\bbtheta}=\bar{\bm{g}}(\bbtheta).
%\end{equation}

For any $\gamma\in [0,1)$, it can be further shown that albeit not symmetric, matrix $\bar{\bm{H}}$ is negative definite, in the sense that $\bm{\theta}^\top \bar{\bm{H}}\bm{\theta}<0$ for any $\bm{\theta}\ne \bm{0}$. Appealing to standard linear systems theory (see e.g., \cite{lyapunov}), we have that the ODE \eqref{eq:ode} admits a globally, asymptotically stable equilibrium point $\bbtheta^*$, dictated by 
\begin{equation}
\bar{\bm{g}}(\bbtheta^*)=\bar{\bm{H}}\bbtheta^*+\bar{\bm{b}}=\bm{0}.
\end{equation}

\subsection{Decentralized Temporal Difference Learning}

The goal of this paper is to investigate the policy evaluation problem in the context of multi-agent reinforcement
learning (MARL), where a group of agents operate to evaluate the value function in an environment.
Suppose there is a set $\mathcal{M}$ of agents with $|\mathcal{M}|=M$, distributed across a network denoted by $\mathcal{G}=(\mathcal{M},\mathcal{E})$, where $\mathcal{E}\subseteq \mathcal{M}\times \mathcal{M}$ represents the edge set. 
 Let $\mathcal{N}_m\subseteq \mathcal{M}$ collect the neighbor(s) of agent $m\in\mathcal{M}$, for all $m\in\mathcal{M}$. We assume that each agent locally implements a stationary policy $\mu_m$. As explained in the centralized setting, when combined with fixed policies $\{\mu_m\}_{m\in\mathcal{M}}$, the multi-agent MDP can be described by the following $6$-tuple
 \begin{equation}
 \label{eq:6tuple}
 \big(\mathcal{S},\{ \mathcal{A}_m\}_{m=1}^M,P,\{R_m\}_{m=1}^M,\gamma,\mathcal{G}\big)
 \end{equation}
 where $\mathcal{S}$ is a finite set of states shared by all agents, $\mathcal{A}_m$ is a finite set of actions available to agent $m$, and $R_m$ is the immediate reward observed by agent $m$. 
It is worth pointing out that, here, we assume
there is no centralized controller that can observe all information; instead, every agent can observe the joint state vector $s\in\mathcal{S}$, yet its action $a_m\in\mathcal{A}_m$ as well as reward $R_m(s,s')$ is kept private from other agents.
 
 Specifically, at time instant $k$, each agent $m$ observes the current state $s(k)\in\mathcal{S}$ and chooses action $a\in\mathcal{A}_m$ according to a stationary policy $\mu_m$. Based on the joint actions of all agents, the system transits to a new state $s(k+1)$, for which an expected local reward $r_m(k)=R_m(s(k),s(k+1))$ is revealed to agent $m$. 
 The objective of multi-agent policy evaluation is to cooperatively compute the average of the expected sums of discounted rewards from a network of agents, given by
 \begin{equation}
 v_{\mathcal{G}}(s)=\mathbb{E}\bigg[\frac{1}{M}\!\sum_{m\in\mathcal{M}}\sum_{k=0}^{\infty} \gamma^k R_m(s(k),s(k+1))\Big| s(0)=s
 \bigg].
 \end{equation}
 Similar to the centralized case, one can show that $v_{\mathcal{G}}(s)$ obeys the following multi-agent Bellman equation 
 \begin{equation}
 	\label{eq:multibe}
 	v_{\mathcal{G}}(s)=\!\sum_{s'\in\mathcal{S}} \!P_{ss'}\!\Big[\frac{1}{M}\!\sum_{m\in\mathcal{M}}\! R_m(s,s') +\gamma v_{\mathcal{G}}(s') \Big],\quad \forall s\in\mathcal{S}.
 \end{equation}
 
 Again, to address the ``curse of dimensionality'' in exact computation of $v_\mathcal{G}$ when the space $\mathcal{S}$ grows large, we are particularly interested in low-dimensional (linear) function approximation $\tilde{v}_{\mathcal{G}}(s)$ of $v_{\mathcal{G}}(s)$ as given in \eqref{eq:linapp0}, or \eqref{eq:linapp1} in a matrix-vector representation.
 
 Define
 $\bm{b}_m(k):=r_m(k)\bm{\phi}(s(k))$, $\bar{\bm b}_m=\mathbb{E}_{\pi}[\bm{b}_m(k)]$, $\bm{b}_{\mathcal{G}}:=\frac{1}{M}\sum_{m\in\mathcal{M}}\bm{b}_m(k)$ and $\bar{\bm{b}}_{\mathcal{G}}:=\frac{1}{M}\sum_{m\in\mathcal{M}} \bar{\bm b}_m
$. As all agents share the same environment by observing a common state vector $s(k)$, and differ only in their rewards, the parameter vector $\bm{\theta}^\ast$ such that the linear function approximator 
 $\bm{\tilde{v}}_{\mathcal{G}}=\bm{\Phi}\bm{\theta}^\ast$ satisfies the multi-agent Bellman equation \eqref{eq:multibe}; that is,
 \begin{equation}
 \label{eq:multitheta}
 \bar{\bm{H}}\bm{\theta}^\ast+\bar{\bm b}_{\mathcal{G}}=\bm{0}
 \end{equation}

We are ready to study a standard consensus-based distributed variant of the centralized TD(0) algorithm, which is tabulated in Algorithm \ref{alg:dtd} for reference.
%
%In a decentralized setting, where a network of agents collaborate to learn the parameter vector $\bbtheta$, by 
%by means of locally updating their own parameter vector $\bbtheta_m$, $m\in\mathcal{M}$ and communicating with their direct neighbors to exchange $\bbtheta_m$.
%Specifically, at time instant $k$, the state transition from $s(k)$ to $s(k+1)$ is public information that all agents can observe; the reward $r_m(k)$, $m\in\mathcal{M}$, however, is private and is observable only for the local agent itself. (For the rest of this paper, 
%In the rest of this paper, without repeating definition, all variables with subscript $m$ represent local variables which vary across agents because of the private reward $r_m$ and local parameter $\bbtheta_m$.) 
Specifically, at the beginning of time instant $k$, each agent $m$ first observes $(s(k),s(k+1),R_m(s(k),s(k+1)))$ and calculates the local gradient
\begin{align}\label{eq:gradientm}
\bm{g}_m(\bbtheta_m(k),\xi_k)&:=\bbphi(s(k))\! \left[\gamma\bbphi^\top(s(k+1))-\bbphi^\top(s(k))\right]\bbtheta_m(k)+r_m(k)\bbphi(s(k))
\end{align}
Upon receiving estimates $\{\bm{\theta}_{m'}(k)\}$ from its neighbors $m'\in\mathcal{N}_m$, agent $m$ updates its local estimate $\bm{\theta}_m(k)$ according to the following recursion
%Upon receiving the information from its neighbors and obtaining the public and private observations, agent $m$ updates $\bbtheta_m$ according to the following decentralized variant of TD(0) learning rule
\begin{equation}\label{eq:detd}
\bbtheta_m(k+1)=\!\!\sum_{m'\in\mathcal{M}}\!\!W_{mm'}\bbtheta_{m'}(k)+\alpha \bm{g}_m(\bbtheta_m(k),\xi_k),\,~ \forall m\in\mathcal{M}
\end{equation}
where $W_{mm'}$ is a weight attached to the edge $(m,m')$; and $W_{mm'}>0$ if $m'\in\mathcal{N}_m$, and $W_{mm'}=0$, otherwise. 
Throughout this paper, we have following assumption on the network.
\begin{As}
The communication network is connected and undirected, and the associated weight matrix $\bm{W}$ is a doubly stochastic matrix. 
\end{As}

%We make the standard assumption that matrix $\bm{W}:=[W_{mm'}]$ is a doubly stochastic matrix.
% \cite{consensus}.

\begin{algorithm}[t]
	\caption{\small {Decentralized TD(0) learning}}\label{alg:dtd} 
	%\hspace*{0.02in} {\bf Input:} 
	\algsetup{linenosize=\footnotesize}
	\small%\footnotesize %, \footnotesize, \scrip'ze, or \tiny
	\begin{algorithmic}[1]
		\STATE \textbf{Input:} stepsize $\alpha>0$, feature matrix $\bm{\Phi}$, and weight matrix $\bm{W}$. 
		\STATE \textbf{Initialize:} $\{\bbtheta_m(0)\}_{m\in\mathcal{M}}$.
		\FOR{$k=0,1,\cdots,K$}
		\FOR{$m=1,2,\cdots, M$}	
		\STATE Agent $m$ receives $\bbtheta_{m'}(k)$ from its neighbors $m'\in\mathcal{N}_m$;
		\STATE Agent $m$ observes $(s(k),s(k+1),r_m(k))$, and computes $\bm{g}_m(\bbtheta_m(k))$ according to \eqref{eq:gradientm};
		\STATE Agent $m$ updates $\bbtheta_m(k)$ via \eqref{eq:detd}, and broadcasts $\bbtheta_m(k+1)$ to its neighbors $m'\in\mathcal{N}_m$.
		\ENDFOR
		\ENDFOR			
	\end{algorithmic}
\end{algorithm}

For ease of exposition, we stack up all local parameter estimates $\{\bm{\theta}_m\}_{m\in\mathcal{M}}$ into matrix 
\begin{equation}
\BTheta:=\begin{bmatrix} \bbtheta_1^\top  \\ \bbtheta_2^\top \\ \vdots  \\ \bbtheta_{M}^\top \end{bmatrix}\in\mathbb{R}^{M\times p}.
\end{equation}
and similarly for all local gradient estimates $\{\bm{g}_m(\bbtheta_m)\}_{m\in\mathcal{M}}$
 \begin{equation}\label{eq:gmatrix}
 \bm{G}(\BTheta,\xi_k):=\begin{bmatrix} \bm{g}_1^\top(\bbtheta_1,\xi_k)  \\ \bm{g}_2^\top(\bbtheta_2,\xi_k) \\ \vdots  \\ \bm{g}_{M}^\top(\bbtheta_{M},\xi_k) \end{bmatrix}\in\mathbb{R}^{M\times p}
 \end{equation}
which admits the following compact representation
{\color{black}
 \begin{equation}
 \bm{G}(\BTheta,\xi_k)=\BTheta \bm{H}^\top(\xi_k)+\bm{r}(k)\bbphi^\top(s(k))
\end{equation}}
where $\bm{r}(k)=[r_1(k)~r_2(k)~\cdots ~r_M(k)]^\top$ concatenates all local rewards.
With the above definitions, the decentralized TD(0) updates in \eqref{eq:detd} over all agents can be collectively re-written as follows
\begin{equation}\label{eq:compdy}
\BTheta(k+1)=\bm{W}\BTheta(k)+\alpha \bm{G}(\BTheta(k),\xi_k).
\end{equation}

In the sequel, we will investigate finite-sample analysis of the decentralized TD(0) learning algorithm in \eqref{eq:compdy} in two steps. 
%Each local parameter converges is not preferred. 
First, we will show that all local parameters reach a consensus, namely, converge to their average. Subsequently, we will prove that the average converges to the Bellman optimum $\bbtheta^*$.
%Instead, we will prove the convergence by showing the convergence of an average system (AS) and a difference system (DS) induced by \eqref{eq:compdy}.

 To this end, let us define the average $\bar{\bbtheta}:=(1/M)\BTheta^T\bm{1}$ of the parameter estimates by all agents, which can be easily shown using \eqref{eq:compdy} to exhibit the following average system (AS) dynamics
\begin{flalign}\label{eq:averdy}
&{\rm AS}: ~~~~~~\bar{\bbtheta}(k+1)=\bar{\bbtheta}(k)+\frac{\alpha}{M}\bm{G}^\top(\BTheta(k),\xi_k)\bm{1}. & 
\end{flalign}
Subtracting from each row of \eqref{eq:compdy} (namely, each local parameter estimate) the average estimate in \eqref{eq:averdy}, yields 
\begin{align}
\BTheta(k+1)-\bm{1}\bar{\bbtheta}^\top(k+1)
&=\bm{W}\BTheta(k)\!-\bm{1}\bar{\bbtheta}^\top(k)\!+\alpha \bm{G}(\BTheta(k),\xi_k)\!-\frac{\alpha}{M}\bm{1}\bm{1}^\top\bm{G}(\BTheta(k),\xi_k)\nonumber
\\
&=\bm{W}\BTheta(k)-\bm{1}\bar{\bbtheta}^\top(k)+\alpha\Big(\bm{I}-\frac{\bm{1}\bm{1}^\top}{M}
\Big)\bm{G}(\bm{\Theta}(k),\xi_k).
\label{eq:diffe}
\end{align}

For notational convenience, we define the network difference operator $\DElta:=\bm{I}-(1/M)\bm{1}\bm{1}^\top$. 
Since $\bm{W}$ is a doubly stochastic matrix, it can be readily shown that $\Delta\BTheta=\BTheta-\bm{1}\bar{\bbtheta}^\top$capturing the difference between local estimates and the global average. After simple algebraic manipulations, 
we deduce that the parameter difference system (DS) evolves as follows
\begin{flalign}\label{eq:diffeb}
&{\rm{DS:}} ~~~\DElta \BTheta(k+1)=\bm{W}\DElta \BTheta(k)+\alpha \DElta \bm{G}(\BTheta(k)). \!\!  & 
\end{flalign}
%where $ \DElta \BTheta $ ($\DElta \bm{G}(\BTheta)$) means the difference between the local parameters (resp. update information) and its global average counterpart.

\section{NON-ASYMPTOTIC PERFORMANCE GUARANTEES}

The goal of this paper is to gain deeper understanding of statistical efficiency of decentralized TD(0) learning algorithms, and investigate their finite-time performance. 
In this direction, we will start off by establishing convergence of the DS in \eqref{eq:diffeb}, that is addressing the consensus among all agents. Formally, we have the following result, whose proof is postponed to Appendix~\ref{append:consensus} for readability.

\begin{Thm}\label{thm:consensus}
Assume that all local rewards are uniformly bounded as $ r_m(k)\in[0,\,r_{\max}]$, $\forall m\in\mathcal{M}$, and the feature vectors $ \bbphi(s)$ have been properly scaled such that $ \|\bbphi(s)\| \le 1$, $\forall s\in\mathcal{S}$. For any deterministic initial guess $\BTheta(0)$ and any constant stepsize $ 0<\alpha\le  
%\frac{1}{2}
 (1-\lambda_2^{\bm{W}})/4$, the parameter estimate difference over the network  at any time instant $k\in\mathbb{N}_+$, satisfies the following 
\begin{equation}\label{eq:diftheta}
\|\DElta \BTheta(k)\|_F \le \! \left(\lambda_2^{\bm{W}}\!+\!2\alpha \right)^k \! 
\|\DElta \BTheta(0)\|_F
+\frac{2\alpha\sqrt{M}r_{\max}}{1-\!\lambda_2^{\bm{W}}}
\end{equation}
where $0<\lambda_2^{\bm{W}}<1$ denotes the second largest eigenvalue of $ \bm{W}$.
\end{Thm}

%Note that for the doubly stochastic matrix $\bm{W}$, its second largest eigenvalue $\lambda_2^{\bm{W}}$ must satisfy $0<\lambda_2^{\bm{W}}<1$. 
Regarding Theorem \ref{thm:consensus}, some remarks come in order. 

To start, it is clear that the smaller $\lambda_2^{\bm{W}}$ is, the faster the convergence is. In practice, it is possible that the operator of the multi-agent system has the freedom to choose the weight matrix $\bm{W}$, so we can optimize the convergence rate by carefully designing $\bm{W}$. 
%However, this goes beyond the scope of this paper; interested readers can refer to e.g., \cite{wmatrix}.
Furthermore, as the number $k$ of updates grows large, the first term on the right-hand-side of \eqref{eq:diftheta} becomes negligible, implying that the parameter estimates of all agents converge to a small neighborhood of the global average $\bar{\bm\theta}(k)$, whose size is proportional to the constant stepsize $\alpha>0$ (multiplied by a certain constant depending solely on the communication network). 
%the above theorem implies that all agents will reach a consensus,  in other words, all local parameters $\bbtheta_m$, $m\in\mathcal{M}$ converge at a linear rate to a neighborhood of the average parameter $\bar{\bbtheta}$. The radius of the neighborhood is the stepsize times a constant, which means the convergence rate and the accuracy trade off with each other by stepsize $\alpha$.  
It is also worth mentioning that the upper bound imposed on the stepsize $ 0<\alpha\le   (1-\lambda_2^{\bm{W}})/4$ is just a sufficient but not necessary condition for convergence. In fact, it can be checked that any stepsize $ 0<\alpha <   (1-\lambda_2^{\bm{W}})/4$ can guarantee exponentially fast consensus of the multi-agents' parameter estimates (up to a small constant error).

So far, we have established the convergence of the DS.  What remains is to show that the global average $\bar{\bbtheta}(k)$ converges to the optimal parameter value $\bbtheta^*$ [cf. \eqref{eq:multitheta}], which is equivalent to showing convergence of the AS in \eqref{eq:averdy}. In this paper, we investigate finite-time performance of decentralized TD(0) learning from 
data samples observed in two different settings, that is the i.i.d. setting as well as the Markovian setting, which occupy the ensuing two subsections.
\subsection{The I.I.D. Setting}

In the i.i.d. setting, we assume that
 data observations $\{(s(k),s(k+1),\{r_m(k)\}_{m\in\mathcal{M}})\}_{k\in\mathbb{N}^+}$ sampled along the trajectory of the underlying Markov chain are i.i.d.. Nevertheless, $s(k)$ and $s(k+1)$ are dependent within each data tuple. 
 Indeed, the i.i.d. setting can be regarded as a special case of the Markovian setting detailed in the next subsection, after the Markov chain has reached a stationary distribution. 
That is, the i.i.d. setting or assuming i.i.d. samples $\{(s(k),s(k+1), r_m(k))\}_k$ is equivalent to considering Markov chains in stationary distributions. To see this, 
consider the probability of the tuple $(s(k),s(k+1), r_m(k))$ taking any value $(s,s',r_m)\subseteq \mathcal{S}\times\mathcal{S}\times \mathbb{R}$
\begin{equation}
{\rm Pr}\!\left\{(s(k),s(k+1))=(s,s')\right\}=\pi(s) P(s,s').
\end{equation}
An alternative way to obtain i.i.d. samples is to generate independently a number of trajectories
and using first-visit methods; see details in \cite{book1996ndp}.  

With i.i.d. data samples, we can establish the following result which
characterizes the relationship between $(1/M)\bm{G}^\top(\BTheta,\xi_j)\bm{1}$ and $\bar{\bm{g}}$. %All the proofs of this paper are provided in the supplementary materials.

\begin{Lem}\label{lem:ubiased}Let $\{\mathcal{F}(k)\}_{k\in\mathbb{N}^+}$ be an increasing family of $\sigma$-fields, with $\BTheta(0)$ being $\mathcal{F}(0)$-measurable, and $\bm{G}(\BTheta(k),\xi_k)$ being $\mathcal{F}(k)$-measurable.
The average $(1/M)\bm{G}^\top(\BTheta(k),
\xi_k)\bm{1}$ of the gradient estimates at all agents
is an unbiased estimate of $\bar{\bm{g}}(\bar{\bbtheta}(k))$; that is,
\begin{equation}\label{eq:unbiased}
\mathbb{E}_{\pi}\!\left[\frac{1}{M}\bm{G}^\top(\BTheta(k),\xi_k)\bm{1}-\bar{\bm{g}}(\bar{\bbtheta}(k))\Big | \mathcal{F}(k) \right]=\bm 0, \forall \xi_k
\end{equation}
and the variance satisfies 
\begin{equation}
\mathbb{E}_{\pi}\!\left[\left\|\frac{1}{M}\bm{G}^\top(\BTheta(k),\xi_k)\bm{1}-\bar{\bm{g}}(\bar{\bbtheta}(k))\right\|^2 \Big | \mathcal{F}(k) \right]\le 4\beta^2 \|\bar{\bbtheta}(k)-\bbtheta^*\|^2+4\beta^2\|\bbtheta^*\|^2+8r_{\max}^2, \forall \xi_j
\end{equation}
where $\beta$ is the maximum spectral radius 
%	(i.e., the largest possible absolute value of eigenvalues of) 
	of matrices $\bm{H}(\xi_k)-\bar{\bm{H}}$ for all $k$.
\end{Lem}

The proof is relegated to Appendix~\ref{append:iidunbiased}.
This lemma suggests that $(1/M)\bm{G}^\top(\BTheta(k),\xi_j)\bm{1}$ is a noisy estimate of $\bar{\bm{g}}(\bar{\bbtheta}(k))$, and the noise is zero-mean and its variance depends only on $\bar{\bbtheta}(k)$. Evidently, the maximum spectral radius of $\bm{H}(\xi_k)-\bar{\bm H}$ can be upper bounded by $ 2(1+\gamma)$ using the definitions of $\bm{H}(\xi_k)$ in \eqref{eq:hmatrix} and $\bar{\bm H}$ in  \eqref{eq:hbar}. 
%and the constants regarding the feature matrix and the MDP.

We are now ready to state our main convergence result in the i.i.d. setting. 
\begin{Thm}\label{thm:coni.i.d.}
Letting $\lambda_{\rm max}^{\bar{\bm{H}}}<0$ denote the largest eigenvalue of $\bar{\bm{H}}$ given in \eqref{eq:hbar}. For any constant stepsize  $0<\alpha\le-\frac{\lambda^{\bar{\bm{H}}}_{\rm max}}{2[4\beta^2+(\lambda^{\bar{\bm{H}}}_{\rm min})^2]}$,  the average parameter estimate over all agents converges linearly to a small neighborhood of the equilibrium point $\bm{\theta}^\ast$; i.e.,
\begin{equation}
\mathbb{E}\!\left[\left\|\bar{\bbtheta}(k)-\bbtheta^*\right\|^2\right] \le c_1^k \left\|\bar{\bbtheta}(0)-\bbtheta^*\right\|^2 + c_2 \alpha
\end{equation}
where the constants $0<c_1:=1+2\alpha \lambda^{\bar{\bm{H}}}_{\rm max}+8\alpha^2\beta^2+2\alpha^2 (\lambda^{\bar{\bm{H}}}_{\rm min})^2<1$ and $c_2:=\frac{8\beta^2 \|\bbtheta^*\|^2+16 r_{\max}^2}{-\lambda^{\bar{\bm{H}}}_{\rm max}}$.
\end{Thm}

Please see a proof in Appendix~\ref{append:iidparaconv}.
Particularly for the i.i.d. setting, the AS drives $\bar{\bbtheta}(k)$ to the optimal solution $\bbtheta^*$ as SGD does, which is indeed due to the fact that $(1/M)\bm{G}^\top(\BTheta(k),\xi_j)\bm{1}$ is an unbiased estimate of $\bar{\bm{g}}(\bar\bbtheta(k))$.

Putting together the convergence result of the global parameter estimate average in Theorem~\ref{thm:coni.i.d.} as, well as, the established consensus among the multi-agents' parameter estimates in Theorem~\ref{thm:consensus},
it follows readily convergence of the local parameter estimates $\{\bm{\theta}_m \}_{m\in\mathcal{M}}$, summarized in the next proposition, for which the proof is provided in  Appendix~\ref{append:iidavecon}.

\begin{Prop}\label{prop:ageni.i.d.pro}
Choosing any constant stepsize 
 $0<\alpha<\alpha_{\max}\triangleq
 \min\Big\{\frac{1-\lambda_2^{\bm{W}}}{4}, - \frac{\lambda^{\bar{\bm{H}}}_{\rm max}}{2[4\beta^2+(\lambda^{\bar{\bm{H}}}_{\rm min})^2]}\Big\}$, then the
 decentralized TD(0) update in \eqref{eq:detd} 
 guarantees that each local parameter estimate $\bm{\theta}_m$ converges linearly to a neighborhood of the optimum $\bbtheta^*$; that is,
\begin{equation}
\mathbb{E}\!\left[\big\|\bbtheta_m(k)-\bbtheta^*\big\|^2\right]\le c_3^kV_0+ c_4\alpha,\quad \forall m\in\mathcal{M}
\end{equation}
 where the constants $c_3:=\max\{(\lambda_2^{\bm{W}}+2\alpha_{\max} )^{2},\,c_1\}$, $V_0:=2\max\{4\|\DElta \BTheta(0)\|_F^2,\, 2\|\bar{\bbtheta}(0)-\bbtheta^*\|^2\}$, and $c_4:=\alpha_{\max}\frac{8 M^2r_{\max}^2}{(1-\lambda_2^{\bm{W}})^2}+\frac{16\beta^2 \|\bbtheta^*\|^2+32 r_{\max}^2}{-\lambda^{\bar{\bm{H}}}_{\rm max}}$.
\end{Prop}

\subsection{The Markovian  Setting}
Although the i.i.d. assumption on the data samples $\{(s(k),s(k+1), r_{m}(t))\}_k$ helps simplify the analysis of TD(0) learning, it represents only an ideal setting, and undermines the practical merits. In this subsection, we will consider a more realistic scenario, where data samples are collected along the trajectory of a single Markov chain starting from any initial distribution. 
%In Markov observation case, the initial state can start anytime rather than after the MDP enters stationary state as the I.I.D case assumes. 
For the resultant Markovian observations, we introduce an important result bounding the bias between the time-averaged `gradient estimate' $\bm{G}(\bm{\Theta},\xi_k)$ and the limit $\bar{\bm{g}}(\bar{\bm{\theta}})$, where $\xi_k$ captures all the randomness corresponding to the $k$-th transition $(s(k),s(k+1),\{r_m(k)\}_{m\in\mathcal{M}})$.  
%here you should first define this $\xi$ formally.... where $\xi_j$, $j\in\mathbb{N}^+$ summarizes the stochasticity during the $j$-th state transition. } 
%, on which the convergence of the average parameter is built.

\begin{Lem}\label{lem:markov}
Let $\{\mathcal{F}(k)\}_{k\in\mathbb{N}^+}$ be an increasing family of $\sigma$-fields, with $\BTheta(0)$ being $\mathcal{F}(0)$-measurable, and $\bm{G}(\BTheta,\xi_k)$ being $\mathcal{F}(k)$-measurable.
Then, for any given $\BTheta\in \mathbb{R}^p$ and any integer $j \in\mathbb{N}^+$, 
%there exists a monotonically decreasing function $\sigma_k(K): \mathbb{N}^{+}\rightarrow \mathbb{R}^+$ that satisfies $\lim_{K\rightarrow \infty}\sigma_k(K)=0$, such that 
the following holds
\begin{align}
&~~~ \Big\|\frac{1}{KM}\!\!\sum_{j=k}^{k+K-1}\!\mathbb{E}\!\left[\bm{G}^\top(\BTheta,\xi_j)\bm{1}\big|\mathcal{F}(k)
%\xi_0\cdots\xi_{i-1}
\right]-\bar{\bm{g}}(\bar{\bbtheta})\Big\|\le \sigma_k(K)(\|\bar{\bbtheta}-\bbtheta^*\|+1).
\label{eq:as1}
\end{align}
where $\sigma_k(K):=\frac{(1+\gamma)\nu_0\rho^k}{(1-\rho)K}\times \max\{ 2\|\bbtheta^*\|+r_{\max}, 1\}$, with constants $\nu_0>0$ and $0<\rho<1$ determined by the Markov chain. In particular for any $k\in \mathbb{N}^+$, it holds that $\sigma_k(K)\le\frac{(1+\gamma)\nu_0}{(1-\rho)K}\times \max\!\big\{ 2\|\bbtheta^*\|+r_{\max}, 1\big\}\triangleq \sigma(K)$.
\end{Lem}

%\gang{The above lemma means that the
%average of $K$ consecutive observations of $(1/M)\bm{G}^\top(\BTheta)\bm{1}$ can approximate $\bar{\bm{g}}(\bar{\bbtheta})$ with bounded error.}
The detailed proof is included in Appendix~\ref{append:marbiased}.
Comparing Lemma \ref{lem:markov} with Lemma \ref{lem:ubiased}, the consequence on the update \eqref{eq:compdy} due to the Markovian observations is elaborated in the following two remarks.
%The key difference between I.I.D and Markov observations as elaborated in the following remark can be made clear by comparing Lemma~\ref{lem:ubiased} and Lemma~\ref{lem:markov}. 

\begin{Remk}
In the Markovian setting, per time instant $k\in\mathbb{N}$, the term $(1/M)\bm{G}^\top(\BTheta(k),\xi_k)\bm{1}$ is a
biased estimate of $\bar{\bm g}(\bar{\bm{\theta}}(k))$, but its time-averaged
bias over a number of future consecutive observations can be upper bounded in terms of the estimation error $\|\bar{\bm \theta}(k)-\bm{\theta}^\ast\|$. Nonetheless, the instantaneous bias, that is when $K=1$, may be sizable or even uncontrollable as there is no constraint on $\sigma(1)$.
\end{Remk}

\begin{Remk}
The results in Lemma \ref{lem:ubiased} for i.i.d. samples correspond to requiring $\sigma(K)=0$ for all $K\in\mathbb{N}^+$ in Lemma \ref{lem:markov}. That is, the i.i.d. setting is indeed a special case of the Markovian one. 
%If $\sigma(K)=0$ holds for any integer $T\ge1$, \eqref{eq:as1} is simplified to \eqref{eq:unbiased}, that is, the Markov observation reduces to I.I.D.
\end{Remk}

In fact, due to the unbiased `gradient' estimates under i.i.d. samples, we were able to directly investigate the convergence of $\bar{\bbtheta}(k)-\bbtheta^*$.
In the Markovian setting however, since we have no control over the instantaneous gradient bias, it becomes challenging, if not impossible, to directly establish convergence of $\bar{\bbtheta}(k)-\bbtheta^*$ as dealt with in the i.i.d. setting. 
In light of the result on the bounded time-averaged gradient bias in Lemma \ref{lem:markov}, 
  we introduce the following multi-step Lyapunov function that 
involves $K$ future consecutive estimates $\{\bar{\bbtheta}(k)\}_{k=k_0}^{k_0+K-1}$:
%\gang{so the proof carries over even if there is no conditioning on the $\sigma$-field $\mathcal{F}(k)$ in the definition of the following Lyapunov function?}
\begin{equation}
\mathbb{V}(k):=\sum_{j=k}^{k+K-1}
%\mathbb{E}\!\left[
\left\|\bar{\bbtheta}(j)-\bbtheta^*\right\|^2,
%\big|\mathcal{F}(k)
%\right],
\quad  k\in\mathbb{N}^+.
\end{equation} 

Concerning the multi-step Lyapunov function, we establish the following result and the proof is relegated to Appendix~\ref{append:vandp}.

\begin{Lem}\label{lem:DLya}
%{\color{black} Let $(1+\gamma)<0$ be the largest absolute value of eigenvalues of $\bm{H}(k)$ for all $k\in\mathbb{N}^+$,} and d
Define the following functions 
\begin{align*}
\Gamma_1(\alpha,K)&=32\alpha^3 K^4(1+2\alpha)^{2K-4}+32K\alpha+8\alpha K^2(1+2\alpha)^{K-2}+4 K \sigma(K)\\
\Gamma_2(\alpha,K)&=\big[32\alpha^3 K^4(1+2\alpha)^{2K-4}+32K\alpha+\alpha K^2(1+2\alpha)^{K-2}\big]\|\bbtheta^*\|^2 \nonumber \\
&\quad +\big[4\alpha^3 K^4(1+2\alpha)^{2K-4}+\frac{1}{2}\alpha K^2(1+2\alpha)^{K-2}+4\alpha K\big]r^2_{\max}+\frac{1}{2} K \sigma(K) 
 \end{align*}
There exists a pair of constants $(\alpha_{\max},\, K_{\mathcal{G}})$ such that $0<1+2\alpha K_{\mathcal{G}} \lambda_{\rm max}^{\bar{\bm{H}}}+\alpha \Gamma_1(\alpha_{\max},K_{\mathcal{G}})<1$ holds for any fixed $\alpha\in(0,\,\alpha_{\rm max})$ and $K= K_{\mathcal{G}}$. 
Moreover, the multi-step Lyapunov function satisfies 
\begin{align}\label{eq:DLya}
\mathbb{E}\!\left[\mathbb{V}(k+1)-\mathbb{V}(k)\big|\mathcal{F}(k)\right] 
&\le \alpha
\big[2 K_{\mathcal{G}} \lambda_{\rm max}^{\bar{\bm{H}}}+\Gamma_1(\alpha_{\max},K_{\mathcal{G}})\big]
%\mathbb{E}\!\left[
\big\|\bar{\bbtheta}(k)-\bbtheta^*\big\|^2
%\big|\mathcal{F}(k)\right]
+\alpha\Gamma_2(\alpha_{\max},K_{\mathcal{G}}).
\end{align}
\end{Lem}

%\gang{Here in the Lemma 3 and the proof that follows, please check if you are missing the signs as $\lambda_{\rm max}^{\bar{\bm{H}}}<0$}

%The proof is relegated to Appendix~\ref{append:vandp}.
Here, we show by construction the existence of a pair $(\alpha_{\rm max},\, K_{\mathcal{G}})$ meeting the conditions on the stepsize. Considering the monotonicity of function $\sigma(K)$, a simple choice for $K_{\mathcal{G}}$ is 
\begin{equation}\label{eq:Tmin}
K_{\mathcal{G}}=\min_{K}\Big\{K \big| \sigma(K)<-\frac{1}{4}\lambda_{\rm max}^{\bar{\bm{H}}}\Big
\}.
\end{equation}
Fixing $K=K_{\mathcal{G}}\ge 1$, it follows that
\begin{equation}
2 K \lambda_{\rm max}^{\bar{\bm{H}}}+\Gamma_1(\alpha,K)
= \Gamma_0(\alpha,K_{\mathcal{G}})
\end{equation}
where $\Gamma_0(\alpha,K_{\mathcal{G}})=32\alpha^3 K_{\mathcal{G}}^6(1+2\alpha)^{2K_{\mathcal{G}}-4}+32\alpha+8\alpha K_{\mathcal{G}}^3(1+2\alpha)^{K_{\mathcal{G}}-2} + K_{\mathcal{G}} \lambda_{\rm max}^{\bar{\bm{H}}}$ can be shown to be monotonically increasing in $\alpha$. Considering further that $\Gamma_0(0,K_{\mathcal{G}})= K_{\mathcal{G}} \lambda_{\rm max}^{\bar{\bm{H}}}< 0$, then there exist a stepsize $\alpha_0$ such that $\Gamma_0(\alpha_0,K_{\mathcal{G}})=\frac{1}{2}K_{\mathcal{G}} \lambda_{\rm max}^{\bar{\bm{H}}}<0$ holds.

Setting now
 \begin{equation}\label{eq:alphamax}
 \alpha_{\rm max}:=\min\bigg\{-\frac{1}{2 K_{\mathcal{G}}\lambda_{\rm max}^{\bar{\bm{H}}}},\;\alpha_0
%   \arg_{\alpha}~ \Gamma_0(\alpha,K_{\mathcal{G}})=\frac{1}{2}K_{\mathcal{G}} \lambda_{\rm max}^{\bar{\bm{H}}} 
\bigg \}
 \end{equation}
 then one can easily check that $0<1+2\alpha K \lambda_{\rm max}^{\bar{\bm{H}}}+\Gamma_1(\alpha,K)\le 1+\frac{1}{2}\alpha K_{\mathcal{G}} \lambda_{\rm max}^{\bar{\bm{H}}}<1$ holds true for any constant stepsize $0<\alpha<\alpha_{\rm max}$. In the remainder of this paper, we will work with $K=K_{\mathcal{G}}$ and $0<\alpha<\alpha_{\rm max}$, 
 yielding
 \begin{align}
\Gamma_0(0,K_{\mathcal{G}})=  K_{\mathcal{G}} \lambda_{\rm max}^{\bar{\bm{H}}} &\le  
%\nonumber\\\Gamma_0(\alpha,\mathcal{K}_{\mathcal{G}})
2 K_{\mathcal{G}} \lambda_{\rm max}^{\bar{\bm{H}}}+\Gamma_1(\alpha,K_{\mathcal{G}})\nonumber\\
 &
 \le \frac{1}{2} K_{\mathcal{G}} \lambda_{\rm max}^{\bar{\bm{H}}}\label{eq:stepbound}
 \end{align}
 where the first inequality uses the fact that $\Gamma_0(\alpha,K_{\mathcal{G}})$ is an increasing function of $\alpha>0$, while the second inequality follows from the definition of $\alpha_0$.

Before presenting the main convergence results in the Markovian setting, we provide a lemma that bounds the multi-step Lyapunov function along the trajectory of a Markov chain. This constitutes a building block for establishing convergence of the averaged parameter estimate.
\begin{Lem}\label{lem:Lyadiff}
The multi-step Lyapunov function is upper bounded as follows
\begin{equation}
\mathbb{V}(k)\le c_5 \big\|\bar{\bbtheta}(k)-\bbtheta^*\big\|^2+c_6 \alpha^2,\quad \forall k\in\mathbb{N}^+ 
\end{equation}
where the constants $c_5$ and $c_6$ are given by
\begin{align*}
c_5&:=\frac{(3+12\alpha_{\rm max}^2\big)^{K_{\mathcal{G}}}-1}{2+3\alpha_{\rm max}^2 }\\
c_6&:= \frac{6(3+\! 12\alpha_{\rm max}^2)\big [ (3+\! 12\alpha_{\max}^2)^{K_{\mathcal{G}}-1}\!-1\big ]\!-6K_{\mathcal{G}}+6}{2+12\alpha_{\rm max}^2} 
 (4 \|\bbtheta^*\|^2+r^2_{\max}\big).
\end{align*}
\end{Lem}

We present the proof in Appendix~\ref{append:vandp2}.
With the above two lemmas, we are now on track to state our convergence result for the averaged parameter estimate, in a Markovian setting.

\begin{Thm}\label{thm:avecon}
 Define constants $c_7:=1+(1/2c_5) \alpha_{\max} K_{\mathcal{G}} \lambda_{\rm max}^{\bar{\bm{H}}}\in(0,1)$, and $c'_8:=\big[16\alpha_{\max}^2 K_{\mathcal{G}}^6 (1+2\alpha_{\max})^{2K_{\mathcal{G}}-4}+32K_{\mathcal{G}}  
+2 K_{\mathcal{G}}^3(1+2\alpha_{\max})^{K_{\mathcal{G}}-2}\big]\|\bbtheta^*\|^2+4 K_{\mathcal{G}}r^2_{\max}-\frac{1}{8} K_{\mathcal{G}} \lambda_{\rm max}^{\bar{\bm{H}}} -\frac{\alpha_{\rm max}c_6}{c_5} K_{\mathcal{G}} \lambda_{\rm max}^{\bar{\bm{H}}}$. Then, fixing any constant stepsize $0<\alpha<\alpha_{\rm max}$ and $K= K_{\mathcal{G}}$ defined in \eqref{eq:Tmin}, the averaged parameter estimate $\bar{\bm \theta}(k)$ converges at a linear rate to a small neighborhood of the equilibrium point $\bbtheta^*$; that is,
\begin{align}
& \mathbb{E}\!\left[\big\|\bar{\bbtheta}(k)-\bbtheta^*\big\|^2\right] \le  c_5c_7^k \big\|\bar{\bbtheta}(0)-\bbtheta^*\big\|^2-\frac{ 2c_5c_8'}{K_{\mathcal{G}} \lambda_{\rm max}^{\bar{\bm{H}}}}\alpha + \min\!\big\{1, \,c_7^{k-k_{\alpha}} \big\}  \Big( \alpha^2 c_6-\frac{ 2c_5c_8'}{K_{\mathcal{G}} \lambda_{\rm max}^{\bar{\bm{H}}}}\Big) \label{eq:convaverage}
\end{align}
where $k_{\alpha}:= \max\{k\in\mathbb{N}^+|\rho^{k}\ge  \alpha\}$.
\end{Thm}

The proof is relegated to Appendix~\ref{append:maravecon}.
As a direct consequence of Theorems \ref{thm:consensus} and \ref{thm:avecon}, our final convergence result on all local parameter estimates comes ready.
\begin{Prop}\label{prop:agentm}
%With $K_{\mathcal{G}}$ and $\alpha_{\rm max}$ the same as in Theorem \ref{thm:avecon},
  Choosing a constant stepsize $0<\alpha<\min\!\big\{\alpha_{\rm max},\,(1-\lambda_2^{\bm{W}})/{4}\big\}$, and any integer $K\ge K_{\mathcal{G}}$,  each local parameter $\bm{\theta}_m(k)$ converges linearly to a neighborhood of the equilibrium point $\bbtheta^*$; that is, the following holds true for each $m\in\mathcal{M}$
\begin{align}
&\mathbb{E}\!\left[\big\|\bbtheta_m(k)-\bbtheta^*\big\|^2\right]\le c_9^k\,V'_0+ \frac{8\alpha^2 M r_{\max}^2}{\big(1-\lambda_2^{\bm{W}}\big)^2} -\frac{ 2c_5c_8'}{K_{\mathcal{G}} \lambda_{\rm max}^{\bar{\bm{H}}}}\alpha \! + \! \min\!\big\{1, \,c_7^{k-k_{\alpha}}\big\}  \Big( \alpha^2 c_6-\frac{ 2c_5c_8'}{K_{\mathcal{G}} \lambda_{\rm max}^{\bar{\bm{H}}}}\Big) \nonumber 
\end{align}
where the constants $c_9:=\max\{(\lambda_2^{\bm{W}}+2\alpha_{\max} )^{2},\,c_7\}$, and $V'_0:=2\max\{4\|\DElta \BTheta(0)\|_F^2,\, 2c_5\|\bar{\bbtheta}(0)-\bbtheta^*\|^2\}$.
\end{Prop}

The proof is similar to that of Proposition~\ref{prop:ageni.i.d.pro}, and hence is omitted here.
Proposition~\ref{prop:agentm} establishes that even in a Markovian setting,  
the local estimates produced by decentralized TD(0) learning converge linearly to a neighborhood of the optimum. Interestingly, different than the i.i.d. case, the size of the neighborhood is characterized in two phases, which correspond to Phase I ($k\le k_{\alpha}$), and Phase II ($k>k_{\alpha}$). In Phase I, the Markov is far from its stationary distribution $\pi$, giving rise to sizable gradient bias in Lemma \ref{lem:markov}, and eventually contributing to a constant-size neighborhood $-{ 2c_5c_8'}/(K_{\mathcal{G}} \lambda_{\rm max}^{\bar{\bm{H}}})$; while, after the Markov chain gets close to $\pi$ in Phase II, confirmed by the geometric mixing property, we are able to have gradient estimates of size-$\mathcal{O}(\alpha)$ bias in Lemma \ref{lem:markov}, and the constant-size neighborhood vanishes with $c_7^{k-k_{\alpha}}$.         

\begin{figure*}[t]
	\hspace{-0.2cm}
	\begin{minipage}[b]{0.35\linewidth} % Èç¹ûÒ»ÐÐ·�
		\centering
		\includegraphics[width=2.1in,height=1.65in]{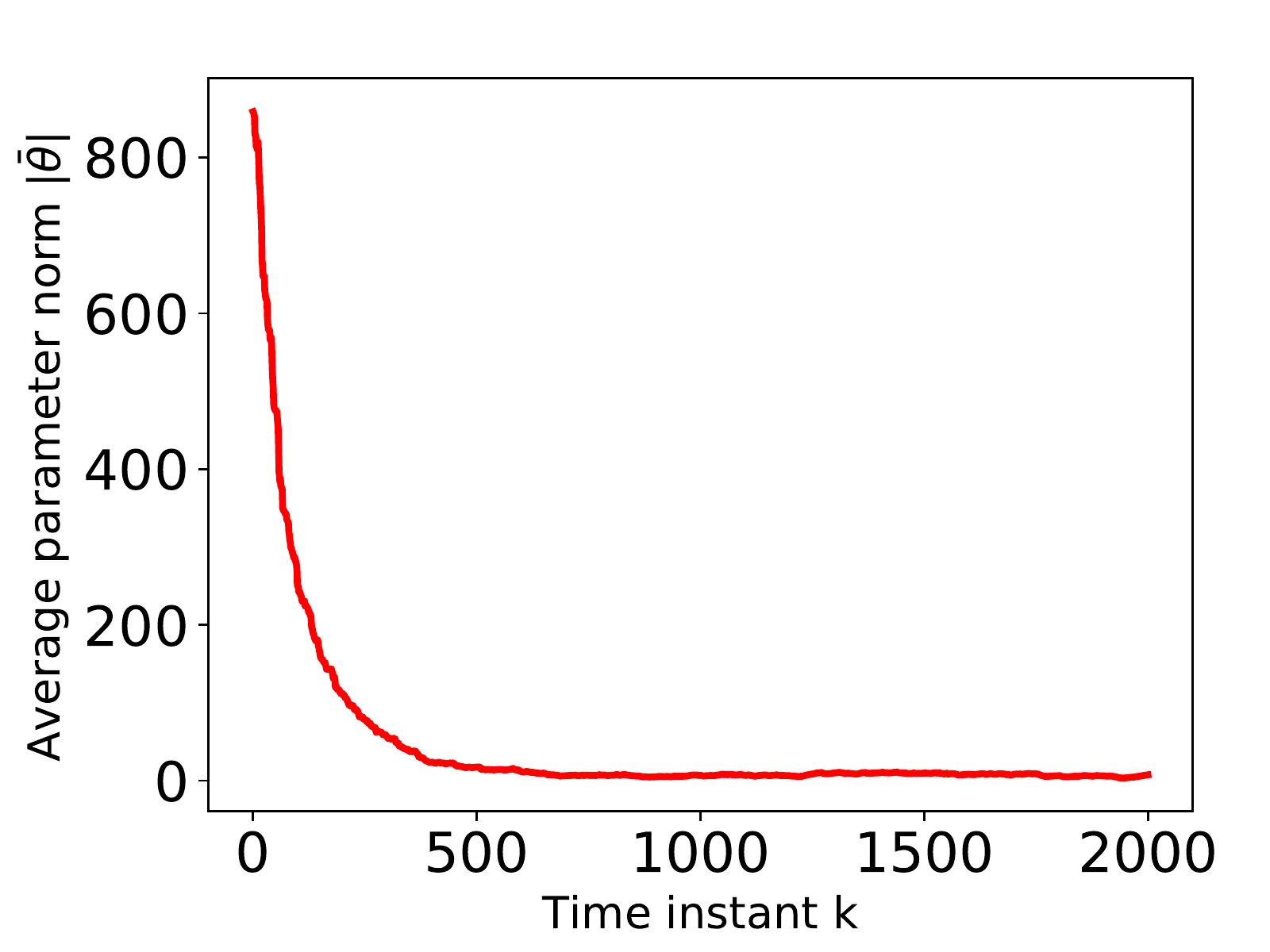}
		%\caption{Loss vs iteration}
		\centerline{(a) Average parameter norm}\medskip
		\label{fig:side:a}
	\end{minipage}%
	\begin{minipage}[b]{0.35\linewidth} % Èç¹ûÒ»ÐÐ·�
		\centering
		\includegraphics[width=2.1in,height=1.65in]{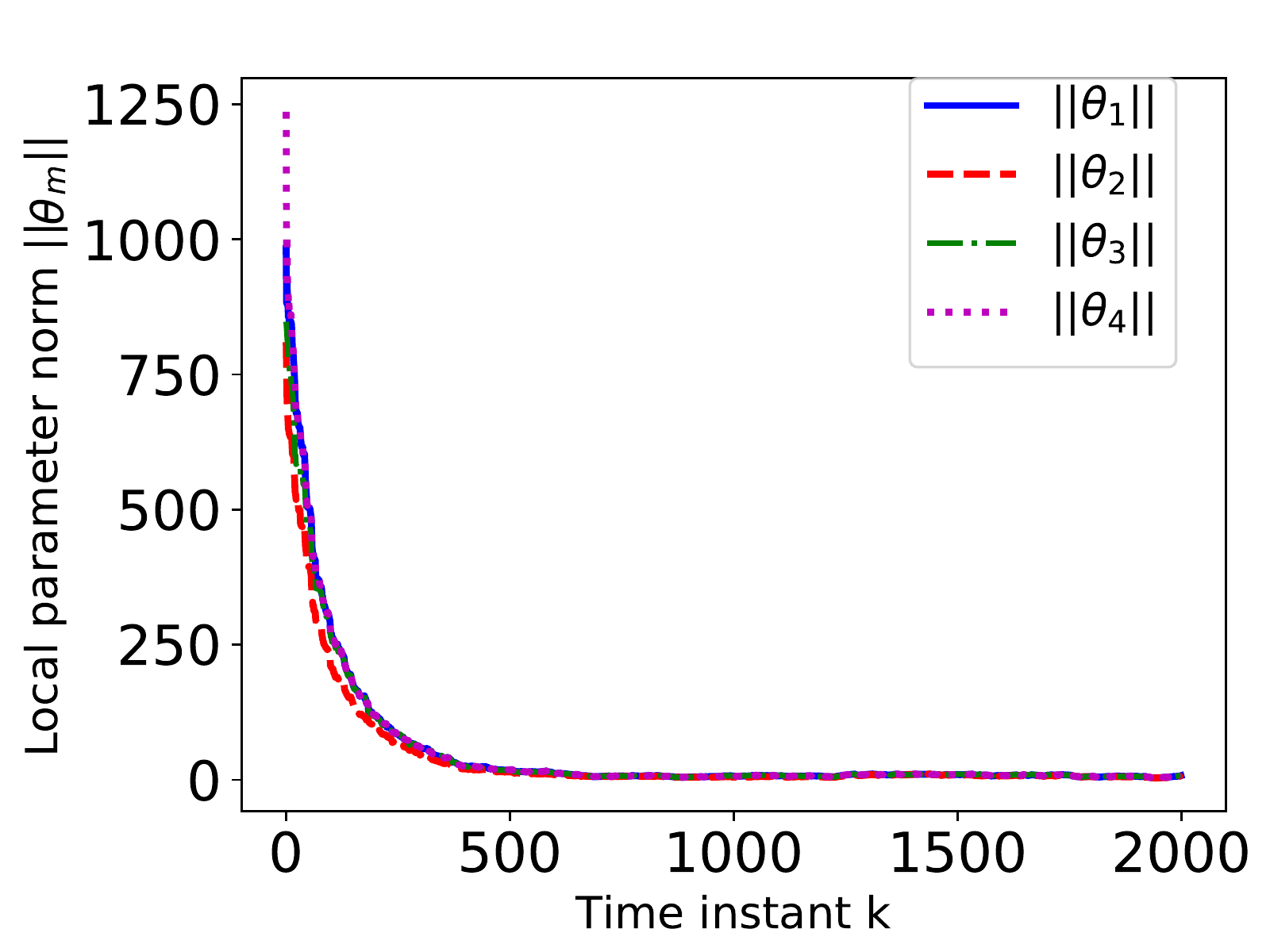}
		%\caption{Small Box}
		\centerline{(b) Local parameters' norm}\medskip
		\label{fig:side:b}
	\end{minipage}%
	\begin{minipage}[b]{0.35\linewidth} % Èç¹ûÒ»ÐÐ·�
		\centering
		\includegraphics[width=2.1in,height=1.65in]{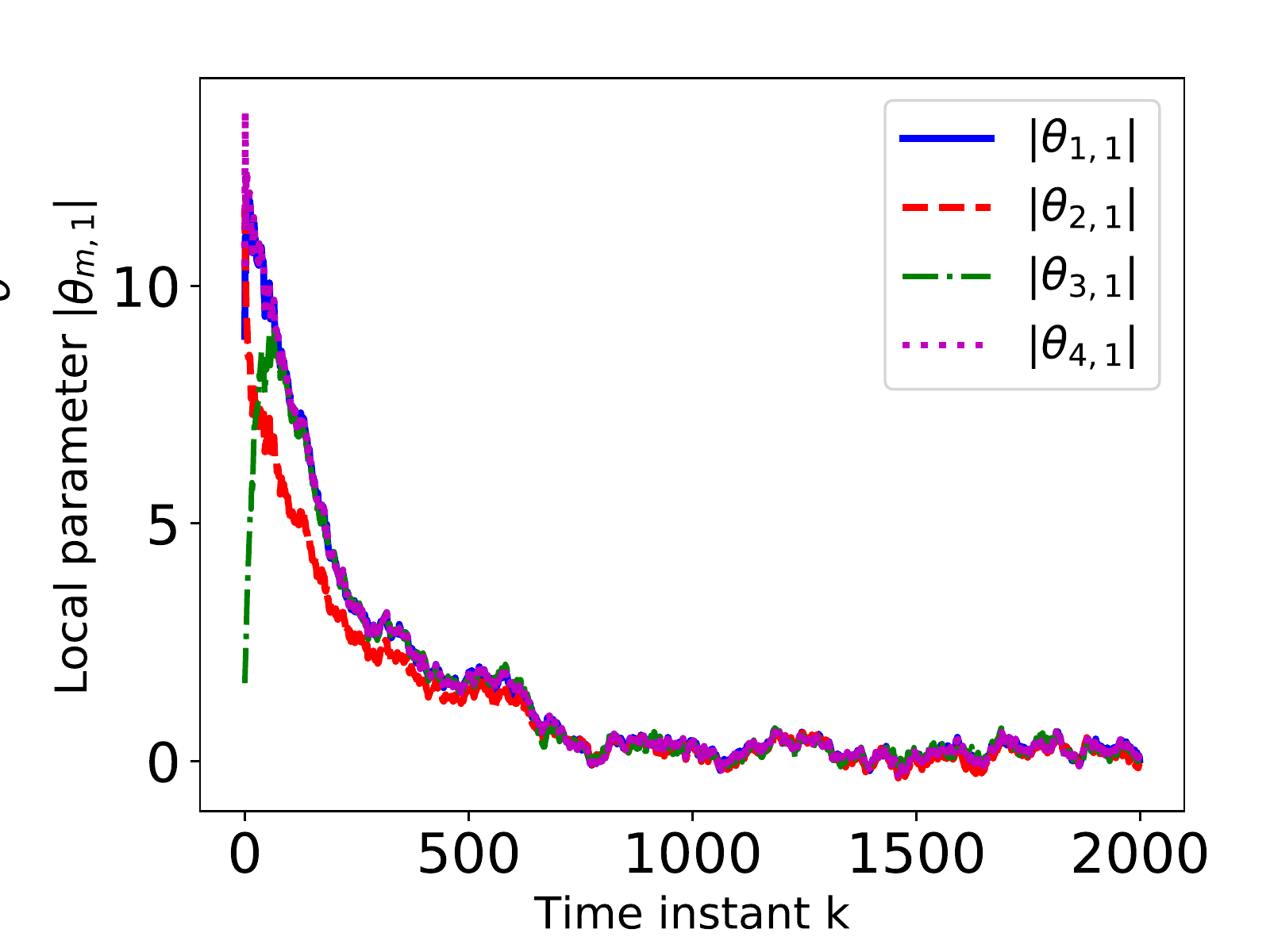}
		\centerline{(c) Local parameters}\medskip
		%\caption{Small Box}
		\label{fig:side:c}
	\end{minipage}%
	\vspace{-0.3cm}
	\caption{Consensus and convergence of decentralized TD(0) learning}\label{fig:candc}
		\vspace{-0.2cm}
\end{figure*}

\section{SIMULATIONS}
In order to verify our analytical results, we carried out experiments on a multi-agent networked system. The details of our experimental setup are as follows: the number of agents $M=30$, the state space size $|\mathcal{S}|=100$ with each state $s$ being a vector of length $|s|=20$, the dimension of learning parameter $\bm{\theta}$ is $p=10$, the reward upper bound $r^{\max}=10$,  and the stepsize $\alpha=0.01$. The feature vectors are cosine functions, that is, $\bm{\phi}(s)=\cos(\bm{A}s)$, where $\bm{A}\in \mathcal{R}^{p\times |s|}$ is a randomly generated matrix. The communication weight matrix $\bm{W}$ depicting the neighborhood of the agents including the topology and the weights was generated randomly, with each agent being associated with $5$ neighbors on average. As illustrated in Fig.~\ref{fig:candc}(a), the parameter average $\bar{\bm{\theta}}$ converges to a small neighborhood of the optimum at a linear rate. To demonstrate the consensus among agents, convergence of the parameter norms $\|\bm{\theta}_m\|$ for $m=1,\,2,\,3,\,4$ is presented in Fig.~\ref{fig:candc}(b), while that of their first elements $|{\theta}_{m,1}|$ is depicted in Fig.~\ref{fig:candc}(c). The simulation results corroborate our theoretical analysis.

\section{CONCLUSIONS}

In this paper, we studied the dynamics of a decentralized linear function approximation variant of the vanilla TD(0) learning, for estimating the value function of a given policy. Allowing for neighboring communications of local parameter estimates, we proved that such decentralized TD(0) algorithms converge linearly to a small neighborhood of the optimum, under both i.i.d. data samples as, well as, the realistic Markovian observations collected along the trajectory of a single Markov chain. 
To address the `gradient bias' in a Markovian setting, our novel approach has been leveraging a carefully designed multi-step Lyapunov function to enable a unique two-phase non-asymptotic convergence analysis. Comparing with previous contributions, this paper provides the first finite-sample error bound for fully decentralized TD(0) learning under challenging Markovian observations. 

%\clearpage
\bibliographystyle{IEEEtranS}
\bibliography{myrP8}

\clearpage
\onecolumn

\begin{center}
	{\Large \bf Supplementary materials for\\ ``Finite-Sample Analysis of Decentralized Temporal-Difference Learning with Linear Function Approximation"}
\end{center}

%\noindent\textbf{Note}. The equations \eqref{eq:obj}—\eqref{eq:ratet} are referenced with respect to the indexing used in the paper.

\appendix

\section{Proof of Theorem ~\ref{thm:consensus}}\label{append:consensus}

\begin{proof}
From the definition of $\bm{G}(\BTheta)$ in \eqref{eq:gmatrix}, we have that
\begin{equation*}
\begin{aligned}
 \bm{G}(\BTheta(k),\xi_k)&
% =\begin{bmatrix} \bm{g}_1(\bbtheta_1(k))^T  \\ \bm{g}_2(\bbtheta_2(k))^T \\ \vdots  \\ \bm{g}_{M}(\bbtheta_{M}(k))^T \end{bmatrix}
 =\begin{bmatrix} \bbtheta_1^\top(k) [\gamma\bbphi(s(k+1))-\bbphi(s(k))]\bbphi^\top(s(k)) \\ \bbtheta_2^\top(k)[\gamma\bbphi(s(k+1))-\bbphi(s(k))]\bbphi^\top(s(k)) \\ \vdots  \\ \bbtheta_M^\top(k)[\gamma\bbphi(s(k+1))-\bbphi(s(k))]\bbphi^\top(s(k)) \end{bmatrix}+\begin{bmatrix} r_1(k)\bbphi^\top(s(k)) \\ r_2(k)\bbphi^\top(s(k)) \\ \vdots  \\ r_M(k)\bbphi^\top(s(k)) \end{bmatrix} \\
 &=\BTheta(k)
 \big[\gamma\bbphi(s(k+1))-\bbphi(s(k))\big]
 \bbphi^\top(s(k))+\bm{r}(k)\bbphi^\top(s(k))\nonumber\\
 &=\BTheta(k) \bm{H}^\top(\xi_k)+\bm{r}(k)\bbphi^\top(s(k))
 \end{aligned}
 \end{equation*}
 where we have used the definitions that  $\bm{r}(k)=[r_1(k)~r_2(k)~\cdots~ r_M(k)]^\top$ and $ \bm{H}(\xi_k):=\bbphi(s(k))[\gamma\bbphi^\top(s(k+1))-\bbphi^\top(s(k))]$. 
Using standard norm inequalties, it follows that
 \begin{align}\label{eq:difnorm}
\left \|\DElta \bm{G}(\BTheta(k),\xi_k)\right\|_F &\le \left\|\left[\gamma\bbphi(s(k+1))-\bbphi(s(k))\right]\bbphi^\top(s(k))\right\|_F\cdot\|\DElta\BTheta(k)\|_F+\left\|\bm{r}(k)\bbphi^\top(s(k))\right\|_F\nonumber\\
 &\le \big[\|\gamma\bbphi(s(k+1))\|_F+\|\bbphi(s(k))\|_F\big]
 \cdot\|\bbphi^\top(s(k))\|_F\cdot\|\DElta\BTheta(k)\|_F+\|\bm{r}(k) \|_F\cdot \|\bbphi(s(k))\|_F\nonumber\\
 &\le (1+\gamma)\|\DElta \BTheta(k)\|_F+\sqrt{M}
 r_{\max}\\
 &\le  2\|\DElta \BTheta(k)\|_F+\sqrt{M}
 r_{\max}
 \end{align}
 where $1+\gamma\le 2$ for the discounting factor $0\le \gamma<1$, 
 and
 the last inequality holds since feature vectors $\|\bbphi(s)\|\le 1$, rewards $r(k)\le r_{\max}$, 
 and the Frobenious norm of rank-one matrices is equivalent to the $\small\ell_2$-norm of vectors. For future reference, notice from the above inequality that $\lambda_{\max}(\bm{H}(\xi_k))\le \|\bm{H}(\xi_k)\|_F=\left\|\left[\gamma\bbphi(s(k+1))-\bbphi(s(k))\right]\bbphi^\top(s(k))\right\|\le 1+\gamma\le 2$, for all $k\in\mathbb{N}^+$.

 It follows from \eqref{eq:diffeb} that
 \begin{align}
 \left\|\DElta \BTheta(k+1)\right\|_F &\le\|\bm{W}\DElta \BTheta(k)\|_F+\alpha\|\DElta \bm{G}(\BTheta(k))\|_F \nonumber\\
 &\le \left[\lambda_2^{\bm{W}}+ 2\alpha \right] \|\DElta \BTheta(k) \|_F+\alpha\sqrt{M}r_{\max} \label{eq:rec}
 \end{align}
where the second inequality is obtained after using \eqref{eq:difnorm}, and the following inequality \cite{wmatrix,admm2019ma}
 \begin{align}
 \|\bm{W}\DElta \BTheta(k)\|_F= \left\|\bm{W}\left(\bm{I}-\frac{1}{M}\bm{1}\bm{1}^\top\right)\bm{\Theta}(k)
\right\|
 \le \lambda_2^{\bm{W}} \|\DElta \BTheta(k)\|_F .
 \end{align}

Then applying \eqref{eq:rec} recursively from iteration $k$ to $0$ gives rise to
\begin{align}
\|\DElta \BTheta(k)\|_F &\le \big (\lambda_2^{\bm{W}}+2\alpha \big)^k \|\DElta \BTheta(0)\|_F+\alpha\sqrt{M}r_{\max}\sum_{i=0}^{k-1} \big (\lambda_2^{\bm{W}}+2\alpha \big)^i \nonumber\\
&\le\big (\lambda_2^{\bm{W}}+2\alpha \big)^k \|\DElta \BTheta(0)\|_F+\frac{\alpha\sqrt{M}r_{\max}}{1-\lambda_2^{\bm{W}}-2\alpha}\nonumber \\
&\le \big (\lambda_2^{\bm{W}}+2\alpha \big)^k \|\DElta \BTheta(0)\|_F+\alpha\cdot\frac{2 \sqrt{M}r_{\max}}{1-\lambda_2^{\bm{W}}}
\end{align}
where the last inequality is a consequence of using the fact that $ 0<\alpha< \frac{1}{2} \cdot \frac{1-\lambda_2^{\bm{W}}}{2}$.  This concludes the proof of Theorem \ref{thm:consensus}. 
\end{proof}

\section{Proof of Lemma ~\ref{lem:ubiased}}\label{append:iidunbiased}
\begin{proof}
Recalling the definitions of $\bm{H}(\xi_k)$ ($\bar{\bm{H}}$) and $\bm{b}({\xi_k})$ ($\bar{\bm{b}}$),
 it is not difficult to verify that in the stationary distribution $\bm{\pi}$ of the Markov chain, the expectations of $\bm{H}(\xi_k)$ and $\bm{b}({\xi_k})$ obey
\begin{equation}
\mathbb{E}_{\pi}[\bm{H}(\xi_k)]=\bar{\bm{H}}
\end{equation}
and
\begin{equation}
\mathbb{E}_{\pi}[\bm{b}_{\mathcal{G}}(\xi_k)]=\bar{\bm{b}}_{\mathcal{G}}.
\end{equation}
Thus,
\begin{equation}
\mathbb{E}_{\pi}\!\left[\frac{1}{M}\bm{G}^\top(\BTheta(k),\xi_k)\bm{1} \Big | \mathcal{F}(k)\right]=\mathbb{E}_{\pi}\big[\bm{H}(\xi_k)\bar{\bbtheta}(k)+\bm{b}_{\mathcal{G}}(\xi_k)\big | \mathcal{F}(k) \big]=\bar{\bm{H}}\bar{\bbtheta}(k)+\bar{\bm{b}}_{\mathcal{G}}
\end{equation}
and its variance satisfies
\begin{align}
\mathbb{E}_{\pi}\!\left[\Big\|\frac{1}{M}\bm{G}^\top(\BTheta(k),\xi_k)\bm{1}-\bar{\bm{g}}(\bar{\bbtheta}(k))\Big\|^2 \Big | \mathcal{F}(k)\right]&=\mathbb{E}_{\pi}\big[\big\|(\bm{H}(\xi_k)-\bar{\bm{H}})\bar{\bbtheta}(k)+\bm{b}_{\mathcal{G}}(\xi_k)-\bar{\bm{b}}_{\mathcal{G}}\big\|^2\big | \mathcal{F}(k) \big] \nonumber\\
&\le \mathbb{E}_{\pi}\big[2\big\|(\bm{H}(\xi_k)-\bar{\bm{H}})\bar{\bbtheta}(k)\big\|^2+2\big\|\bm{b}_{\mathcal{G}}(\xi_k)-\bar{\bm{b}}_{\mathcal{G}}\big\|^2\big | \mathcal{F}(k)\big] \nonumber\\
&\le 2\beta^2 \|\bar{\bbtheta}(k)-\bbtheta^*+\bbtheta^*\|^2+8r_{\max}^2 \nonumber\\
&\le 4\beta^2 \|\bar{\bbtheta}(k)-\bbtheta^*\|^2+4\beta^2\|\bbtheta^*\|^2+8r_{\max}^2
\end{align}
where $\beta$ denotes the largest absolute value of eigenvalues of $\bm{H}(\xi_k)-\bar{\bm{H}}$, for any $k\in\mathbb{N}^+$.
\end{proof}

\section{Proof of Theorem ~\ref{thm:coni.i.d.}}\label{append:iidparaconv}
\begin{proof}
	Clearly, it holds that
\begin{align}
\mathbb{E}_{\pi}[\|\bar{\bbtheta}(k+1)-\bbtheta^*\|^2\big|\mathcal{F}(k)] &=\mathbb{E}_{\pi}\Big[\Big\|\bar{\bbtheta}(k)-\bbtheta^*+\alpha \frac{1}{M}\bm{G}^\top(\BTheta,\xi_k)\bm{1}\Big\|^2\Big|\mathcal{F}(k)\Big]\nonumber\\
&\le \|\bar{\bbtheta}(k)-\bbtheta^*\|^2+2\alpha \left\langle \bar{\bbtheta}(k)-\bbtheta^*, \mathbb{E}_{\pi}\Big[\frac{1}{M}\bm{G}(\BTheta(k),\xi_k)^T\bm{1}\Big | \mathcal{F}(k)\Big] \right\rangle\nonumber\\
&\quad +\alpha^2 \mathbb{E}_{\pi}\Big[\Big\|\frac{1}{M}\bm{G}(\BTheta(k),\xi_k)^T\bm{1}-\bar{\bm{g}}(\bar{\bbtheta}(k))+\bar{\bm{g}}(\bar{\bbtheta}(k))\Big\|^2\big|\mathcal{F}(k)\Big]\nonumber\\
&\le \|\bar{\bbtheta}(k)-\bbtheta^*\|^2+2\alpha \left\langle \bar{\bbtheta}(k)-\bbtheta^*, \bar{\bm{g}}(\bar{\bbtheta}(k))-\bar{\bm{g}}(\bar{\bbtheta}^*)\right\rangle
\nonumber\\
&\quad+2\alpha^2 (\beta^2\|\bar{\bbtheta}\|^2+r_{\max}^2)+2\alpha^2 \|\bar{\bm{g}}(\bar{\bbtheta}(k))-\bar{\bm{g}}(\bar{\bbtheta}^*)\|^2 \nonumber\\
&\le \|\bar{\bbtheta}(k)-\bbtheta^*\|^2+2\alpha \left\langle \bar{\bbtheta}(k)-\bbtheta^*,\bar{\bm{H}}(\bar{\bbtheta}(k)-\bbtheta^*) \right\rangle
\nonumber\\
&\quad+2\alpha^2 (4\beta^2 \|\bar{\bbtheta}-\bbtheta^*\|^2+4\beta^2\|\bbtheta^*\|^2+8r_{\max}^2)+ 2\alpha^2 \|\bar{\bm{H}}(\bar{\bbtheta}(k)-\bbtheta^*) \|^2 \nonumber\\
&\le \left[1+2\alpha \lambda^{\bar{\bm{H}}}_{\rm max}+8\alpha^2\beta^2+2\alpha^2 (\lambda^{\bar{\bm{H}}}_{\rm min})^2 \right] \|\bar{\bbtheta}(k)-\bbtheta^*\|^2\nonumber\\
&\quad +(8\alpha^2\beta^2 \|\bbtheta^*\|^2+16\alpha^2 r_{\max}^2).\label{eq:difone}
\end{align} 
where $\lambda_{\rm max}^{\bar{\bm{H}}}$ and $\lambda^{\bar{\bm{H}}}_{\rm min}$ are the largest and the smallest eigenvalues of $\bar{\bm{H}}$, respectively. Because $\bar{\bm{H}}$ is a negative definite matrix, then it follows that $\lambda^{\bar{\bm{H}}}_{\rm min}<\lambda_{\rm max}^{\bar{\bm{H}}}<0$.

Defining constants $c_1:=1+2\alpha \lambda^{\bar{\bm{H}}}_{\rm max}+8\alpha^2\beta^2+2\alpha^2 (\lambda^{\bar{\bm{H}}}_{\rm min})^2$, and choosing any constant stepsize $\alpha$ obeying $0<\alpha\le-\frac{1}{2}\cdot \frac{\lambda^{\bar{\bm{H}}}_{\rm max}}{4\beta^2+(\lambda^{\bar{\bm{H}}}_{\rm min})^2}$, then we have $c_1<1$ and $\frac{1}{1-c_1}\le
{\color{blue}-}
\frac{1}{\alpha\lambda^{\bar{\bm{H}}}_{\rm max}}$. Now, taking expectation with respect to $\mathcal{F}(k)$ in \eqref{eq:difone} gives rise to
\begin{equation}
\mathbb{E}
\!\left[\|\bar{\bbtheta}(k+1)-\bbtheta^*\|^2 \right]\le c_1 \mathbb{E}\!\left[\|\bar{\bbtheta}(k)-\bbtheta^*\|^2 \right]+(8\alpha^2\beta^2 \|\bbtheta^*\|^2+16\alpha^2 r_{\max}^2).
\end{equation}
Applying the above recursion from iteration $k$ to iteration $0$ yields
\begin{align}
\mathbb{E}\!\left[\|\bar{\bbtheta}(k)-\bbtheta^*\|^2\right] &\le c_1^k \|\bar{\bbtheta}(0)-\bbtheta^*\|^2 +\frac{1-c_1^k}{1-c_1}\left(8\alpha^2\beta^2 \|\bbtheta^*\|^2+16\alpha^2 r_{\max}^2\right) \nonumber\\
&\le c_1^k \|\bar{\bbtheta}(0)-\bbtheta^*\|^2+\frac{8\alpha^2\beta^2 \|\bbtheta^*\|^2+16\alpha^2 r_{\max}^2}{-\alpha\lambda^{\bar{\bm{H}}}_{\rm max}}\nonumber\\
&\le c_1^k \|\bar{\bbtheta}(0)-\bbtheta^*\|^2+\alpha c_2\label{eq:cai.i.d.}
\end{align}
where $c_2:=\frac{8\beta^2 \|\bbtheta^*\|^2+16 r_{\max}^2}{-\lambda^{\bar{\bm{H}}}_{\rm max}}$, and this concludes the proof.
\end{proof}

\section{Proof of Proposition~\ref{prop:ageni.i.d.pro}}\label{append:iidavecon}

\begin{proof}
	We have that
\begin{align}
\mathbb{E}\!\left[\|\bbtheta_m(k)-\bbtheta^*\|^2\right]&=\mathbb{E}\!\left[\|\bbtheta_m(k)-\bar{\bbtheta}(k)+\bar{\bbtheta}(k)-\bbtheta^*\|^2\right]\nonumber \\
&\le 2\mathbb{E}\!\left[\|\bbtheta_m(k)-\bar{\bbtheta}(k)\|^2\right]+2\mathbb{E}\!\left[\|\bar{\bbtheta}(k)-\bbtheta^*\|^2\right] \nonumber\\
&\le 2\mathbb{E}\!\left[\|\DElta \BTheta(k)\|_F^2\right]+2\mathbb{E}\!\left[\|\bar{\bbtheta}(k)-\bbtheta^*\|^2\right]\nonumber\\
&\le  2\mathbb{E}\!\left[\big (\lambda_2^{\bm{W}}+2\alpha \big)^k \|\DElta \BTheta(0)\|_F+\frac{2\alpha\sqrt{M}r_{\max}}{1-\lambda_2^{\bm{W}}}\right]^2+2c_1^k \|\bar{\bbtheta}(0)-\bbtheta^*\|^2 +2\alpha c_2
\nonumber\\
&\le 4\!\big (\lambda_2^{\bm{W}}+2\alpha \big)^{2k} \|\DElta \BTheta(0)\|_F^2+\frac{8\alpha^2 Mr_{\max}^2}{(1-\lambda_2^{\bm{W}})^2}+2c_1^k \|\bar{\bbtheta}(0)-\bbtheta^*\|^2 +2\alpha c_2
.\label{eq:finalconvi.i.d.}
\end{align}
where the third inequality follows from using \eqref{eq:diftheta} and \eqref{eq:cai.i.d.}. Letting $c_3:=\max\{\big (\lambda_2^{\bm{W}}+2\alpha \big)^2,\,c_1\}$, $V_0:=2\max\{4\|\DElta \BTheta(0)\|_F^2,\, 2\|\bar{\bbtheta}(0)-\bbtheta^*\|^2\}$, and $c_4:=\alpha\cdot\frac{8 Mr_{\max}^2}{(1-\lambda_2^{\bm{W}})^2}+\frac{16\beta^2 \|\bbtheta^*\|^2+32 r_{\max}^2}{-\lambda^{\bar{\bm{H}}}_{\rm max}}$, then it is straightforward from \eqref{eq:finalconvi.i.d.} that our desired result follows; that is,
\begin{equation}
\mathbb{E}\!\left[\left\|\bbtheta_m(k)-\bbtheta^*\right\|^2\right]\le c_3^kV_0+ c_4 \alpha
\end{equation}
which concludes the proof.
\end{proof}

\section{Proof of Lemma~\ref{lem:markov}}\label{append:marbiased}
\begin{proof} For notational brevity, let $r_{\mathcal{G}}(k):=(1/M)\sum_{m\in\mathcal{M}} r_m(k)$ for each $k\in\mathbb{N}^+$. It then follows that
\begin{align}
&~~~ \Big\|\frac{1}{KM}\!\!\sum_{j=k}^{k+K-1}\!\mathbb{E}\big[\bm{G}^\top(\BTheta,\xi_j)\bm{1}\big|\mathcal{F}(k)
\big]-\bar{\bm{g}}(\bar{\bbtheta})\Big\|\nonumber\\
&= \Big\| \frac{1}{K}\sum_{j=k}^{k+K-1}\!\mathbb{E}\Big[ \bbphi(s(k))[\gamma\bbphi(s(k+1))-\bbphi(s(k))]^{\top}\bar{\bbtheta}+\frac{1}{M}\bbphi(s(k))\bm{r}^{\top}(k)\bm{1}\Big]-\mathbb{E}_{\pi}\big[ \bm{g}(\bar{\bbtheta})\big] \Big\|\nonumber \\
&= \Big\| \frac{1}{K}\sum_{j=k}^{k+K-1}\sum_{s\in\mathcal{S}} \Big( {\rm Pr}\big[s(j)=s|\mathcal{F}(k)\big]-\pi(s)\Big) \left[\bbphi(s) \big(\gamma P(s,s') \bbphi(s')-\bbphi(s)\big)^{\top}(\bar{\bbtheta}+\bbtheta^*)+r_{\mathcal{G}}(s)\bbphi(s) \right]\Big\| \nonumber \\
&\le \max_{s,s'} \Big\|  \bbphi(s) \big[\gamma P(s,s') \bbphi(s')-\bbphi(s)\big]^{\top}(\bar{\bbtheta}+\bbtheta^*)+r_{\mathcal{G}}(s)\bbphi(s) \Big \|\nonumber \\
&\quad \times \frac{1}{K}\sum_{j=k}^{k+K-1}\sum_{s\in\mathcal{S}} \Big| {\rm Pr}[s(j)=s|\mathcal{F}(k)]-\pi(s)\Big|\nonumber \\
&\le (1+\gamma)\big(\|\bar{\bbtheta}-\bbtheta^*\|+2\|\bbtheta^*\|+r_{\max}\big) \times \frac{1}{K}\sum_{j=k}^{k+K-1} \nu_0 \rho^k \cdot \rho^{j-k}\nonumber \\
&\le \frac{(1+\gamma)\nu_0\rho^k}{(1-\rho)K} (\|\bar{\bbtheta}-\bbtheta^*\|+2\|\bbtheta^*\|+r_{\max})\nonumber\\
&\le \sigma_k(K) \big(\|\bar{\bbtheta}-\bbtheta^*\|+1\big)
\end{align}
where $\sigma_k(K)=\frac{(1+\gamma)\nu_0\rho^k}{(1-\rho)K}\times \max\!\big\{ 2\|\bbtheta^*\|+r_{\max}, 1\big\}$, 
%if $2\|\bbtheta^*\|+r_{\max}>1$, otherwise $\sigma(K)=\frac{(1+\gamma)\mu}{(1-\rho)K}$; 
and the second inequality arises from the fact that any finite-state, irreducible, and aperiodic Markov chains converges geometrically fast (with some initial constant $\nu_0>0$ and rate $0<\rho<1$) to its unique stationary distribution \cite[Thm. 4.9]{mcbook}. Thus, we conclude that Lemma \ref{lem:markov} holds true with monotonically decreasing function $\sigma(K)$ of $K\in\mathbb{N}^+$ as defined above.
\end{proof}

\section{Proof of Lemma~\ref{lem:DLya}}\label{append:vandp}
\begin{proof}
Recalling the definition of our multi-step Lyapunov function, we obtain that
\begin{equation}
\mathbb{E}\!\left[\mathbb{V}(k+1)-\mathbb{V}(k) \big|\mathcal{F}(k)\right]=\mathbb{E}\!\left[\|\bar{\bbtheta}(k+K)-\bbtheta^*\|^2-\|\bar{\bbtheta}(k)-\bbtheta^*\|^2\big|\mathcal{F}(k)\right].
\end{equation}
Thus, we should next derive the bound of the right hand side of above equation.
Following from iterate \eqref{eq:averdy}, we can write
\begin{equation}
\bar{\bbtheta}(k+K)=\bar{\bbtheta}(k)+\frac{\alpha}{M}\sum_{j=k}^{k+K-1}\bm{G}^{\top}(\bm\BTheta(j),\xi_j)\bm{1}.
\end{equation}
As a consequence (without particular statement, the expectation in the rest of this proof is taken with respect to the $\xi_k$ to $\xi_{k+K-1}$ conditioned on $\xi_0$ to $\xi_{k-1}$),
\begin{align}
&\mathbb{E}\Big[\|\bar{\bbtheta}(k+K)-\bbtheta^*\|^2\big|\mathcal{F}(k)\Big]=\mathbb{E}\bigg[ \Big\|\bar{\bbtheta}(k)-\bbtheta^*+\frac{\alpha}{M}\sum_{j=k}^{k+K-1}\bm{G}^{\top}(\bm\BTheta(j),\xi_j)\bm{1}\Big\|^2 \big|\mathcal{F}(k)\bigg]\nonumber\\
&=\mathbb{E}\bigg[ \Big \|\bar{\bbtheta}(k)-\bbtheta^*+\frac{\alpha}{M}\sum_{j=k}^{k+K-1}\Big[\bm{G}^\top(\bm\BTheta(j),\xi_j)\bm{1}-\bm{G}^\top(\bm\BTheta(k),\xi_j)\bm{1}+\bm{G}^\top(\bm\BTheta(k),\xi_j)\bm{1}\Big] \Big \|^2 \Big|\mathcal{F}(k)\bigg] \nonumber\\
&= \|\bar{\bbtheta}(k)-\bbtheta^*\|^2\nonumber
\\&
\quad +2\alpha \mathbb{E}\!\bigg
[ \bigg\langle \bar{\bbtheta}(k)-\bbtheta^*,K\bar{\bm{g}}(\bar{\bbtheta}(k))\!+\!\frac{1}{M}\!\sum_{j=k}^{k+K-1}\!\Big[\bm{G}^\top(\bm\BTheta(j),\xi_j)\bm{1}\!-\!\bm{G}^\top(\bm\BTheta(k),\xi_j)\bm{1}\!+\!\bm{G}^\top(\bm\BTheta(k),\xi_j)\bm{1}\Big]\!-\!K\bar{\bm{g}}(\bar{\bbtheta}(k)) \bigg\rangle\Big|\mathcal{F}(k)\bigg] \nonumber\\
&\quad+\alpha^2\mathbb{E}\bigg[\Big\|\frac{1}{M}\sum_{j=k}^{k+K-1}\Big[\bm{G}^\top(\bm\BTheta(j),\xi_j)\bm{1}-\bm{G}^\top(\bm\BTheta(k),\xi_j)\bm{1}+\bm{G}^\top(\bm\BTheta(k),\xi_j)\bm{1}\Big]\Big\|^2\Big|\mathcal{F}(k)\bigg] \nonumber\\
&= \|\bar{\bbtheta}(k)-\bbtheta^*\|^2+\underbrace{2\alpha \mathbb{E}\!\left[\left\langle \bar{\bbtheta}(k)-\bbtheta^*,K\bar{\bm{g}}(\bar{\bbtheta}(k))-K\bar{\bm{g}}(\bbtheta^*)\right\rangle\Big|\mathcal{F}(k)\right]}_{\text{the~second~term}}\nonumber\\
&\quad+\underbrace{2\alpha \mathbb{E}\bigg[\left\langle \bar{\bbtheta}(k)-\bbtheta^*,\sum_{j=k}^{k+K-1}\frac{1}{M}\Big[\bm{G}^\top(\bm\BTheta(j),\xi_j)\bm{1}-\bm{G}^\top(\bm\BTheta(k),\xi_j)\bm{1}\Big]\right\rangle\Big|\mathcal{F}(k)\bigg]}_{\text{the~third~term}} \nonumber\\
&\quad+ \underbrace {2\alpha \mathbb{E}\bigg[ \left\langle \bar{\bbtheta}(k)-\bbtheta^*,\sum_{j=k}^{k+K-1}\frac{1}{M}\bm{G}^\top(\bm\BTheta(k),\xi_j)\bm{1}-K\bar{\bm{g}}(\bar{\bbtheta}) \right\rangle\Big|\mathcal{F}(k)\bigg]}_{\text{the~fourth~term}} \nonumber\\
&\quad+ \underbrace{\alpha^2\mathbb{E}\bigg[\bigg\|\frac{1}{M}\sum_{j=k}^{k+K-1}\Big[\bm{G}^\top(\bm\BTheta(j),\xi_j)\bm{1}-\bm{G}^\top(\bm\BTheta(k),\xi_j)\bm{1}+\bm{G}^\top(\bm\BTheta(k),\xi_j)\bm{1}\Big]\bigg\|^2\Big|\mathcal{F}(k)\bigg]} \label{eq:thetade}_{\text{the~last~term}}
\end{align}
where the second and the third equality result from adding and subtracting the same terms and the last equality holds since $\bar{g}(\bbtheta^*)=0$. In the following, we will bound the four terms in the above equality.

\begin{itemize}
\item[\textbf{1)}] \textbf{Bounding the second term.}
As a direct result of the definition of $\bar{g}(\bbtheta)$, we have that $\bar{\bm{g}}(\bar{\bbtheta})-\bar{\bm{g}}(\bbtheta^*)=\bar{\bm{H}}(\bar{\bbtheta}-\bbtheta^*)$. Therefore, it holds that
\begin{align}
2\alpha \mathbb{E}\!\left[\left\langle \bar{\bbtheta}(k)-\bbtheta^*,K\bar{\bm{g}}(\bar{\bbtheta}(k))-K\bar{\bm{g}}(\bbtheta^*)\right\rangle\Big|\mathcal{F}(k)\right]&=2\alpha K \mathbb{E} \left[(\bar{\bbtheta}(k)-\bbtheta^*)^{\top}\bar{\bm{H}}(\bar{\bbtheta}(k)-\bbtheta^*) |\mathcal{F}(k)\right]\nonumber\\
&\le 2\alpha K \lambda_{\rm max}^{\bar{\bm{H}}}\|\bar{\bbtheta}(k)-\bbtheta^*\|^2\label{eq:term2}
\end{align}
where $\lambda_{\rm max}^{\bar{\bm{H}}}$ is the largest eigenvalue of $\bar{\bm{H}}$. Because $\bar{\bm{H}}$ is a negative definite matrix, it holds that $\lambda_{\rm max}^{\bar{\bm{H}}}<0$.

\item[\textbf{2)}] \textbf{Bounding the third term.}
Defining first $\bm{p}(k,\BTheta(k),K):=\sum_{j=k}^{k+K-1}\frac{1}{M}\Big[\bm{G}^\top(\bm\BTheta(j),\xi_j)\bm{1}-\bm{G}^\top(\bm\BTheta(k),\xi_j)\bm{1}\Big]$, then it follows that
\begin{align*}
\!\!\!\! \bm{p}(k,\BTheta(k),K)&=\sum_{j=k}^{k+K-2}\frac{1}{M}\left[\bm{G}^\top(\bm\BTheta(j),\xi_j)\bm{1}-\bm{G}^\top(\bm\BTheta(k),\xi_j)\bm{1} \right]\\
&~~~~+\frac{1}{M} \Big[\bm{G}^{\top}(\BTheta(k+K-1),\xi_{k+K-1})\bm{1}-\bm{G}^{\top}(\BTheta(k),\xi_{k+K-1})\bm{1} \Big]\\
&=\bm{p}(k,\BTheta(k),K-1)+\frac{1}{M}\Big[\bm{G}^{\top}(\bm\BTheta(k+K-1),\xi_{k+K-1})\bm{1}-\bm{G}^{\top}(\bm\BTheta(k),\xi_{k+K-1})\bm{1} \Big] \\
&=\bm{p}(k,\BTheta(k),K-1)+\bm{H}(k+K-1)[\bar{\bbtheta}(k+K-1)-\bar{\bbtheta}(k)].
\end{align*}

Recalling that $2$ is the largest absolute value of eigenvalues of $\bm{H}(k)$ for any $k\in\mathbb{N}^+$ (which clearly exists and is bounded due to the bounded feature vectors $\bm{\phi}(s)$ for any $s\in\mathcal{S}$), the norm of $\bm{p}(k,\BTheta(k),K)$ can be bounded as follows
{\color{black}
\begin{align*}
\|\bm{p}(k,\BTheta(k),K)\| &\le\|\bm{p}(k,\BTheta(k),K-1\|+2 \|\bar{\bbtheta}(k+K-1)-\bar{\bbtheta}(k)\|\\
&=\|\bm{p}(k,\BTheta(k),K-1)\|+
2\alpha\, \bigg\|\sum_{j=k}^{k+K-2}\frac{1}{M} \Big[\bm{G}^\top(\bm\BTheta(j),\xi_j)\bm{1}-\bm{G}^\top(\bm\BTheta(k),\xi_j)\bm{1}\Big]\\
&\quad +\sum_{j=k}^{k+K-2}\frac{1}{M}\bm{G}^\top(\bm\BTheta(k),\xi_j)\bm{1} \bigg \| \\
&\le (1+2\alpha)\|\bm{p}(k,\BTheta(k),K-1)\|+2\sum_{j=k}^{k+K-2}\alpha \|\bm{H}(j)\bar{\bbtheta}(k)+b_{\mathcal{G}}\| \\
&\le  (1+2\alpha)\|\bm{p}(k,\BTheta(k),K-1)\|+4\alpha \bigg(\sum_{j=k}^{k+K-2}\|\bar{\bbtheta}(k)\|+\frac{ r_{\max}}{2} \bigg)
\end{align*}}
where the last inequality follows from $\|\bm{H}(j)\bar{\bbtheta}(k)\|\le 2 \|\bar{\bbtheta}(k)\|$ for any $j\ge 0$. Following the above recursion, we can write
\begin{equation}\label{eq:normp}
\begin{aligned}
\|\bm{p}(k,\BTheta(k),K)\| &\le(1+2\alpha)^K \|\bm{p}(k,\BTheta(k),0)\|+4\alpha K \|\bar{\bbtheta}(k)\| \sum_{j=0}^{K-1}(1+2\alpha)^j(K-1-j) \\
&\le 4\alpha  (\|\bar{\bbtheta}(k)\|+\frac{r_{\max}}{2}) \sum_{j=0}^{K-1}(1+2\alpha)^j(K-1-j) 
\end{aligned}
\end{equation}
where the second inequality because $\|\bm{p}(k,\BTheta(k),0)\|=0$.

For any positive constant $x\neq 1$ and $K\in\mathbb{N}^+$, the following equality holds 
\begin{equation}\label{eq:sum}
\sum_{j=0}^{K-1}x^j(K-1-j)=\frac{x^K-Kx+K-1}{(1-x)^2}.
\end{equation}
Substituting $x=(1+2\alpha)$ into \eqref{eq:sum} along with plugging the result into \eqref{eq:normp} yields 
\begin{equation}\label{eq:normpa}
\|\bm{p}(k,\BTheta(k),K)\| \le \frac{(1+2\alpha)^K-2K\alpha-1}{\alpha} K\|\bar{\bbtheta}(k)\|.
\end{equation}

According to the mid-value theorem, there exists some suitable constant $\delta\in[0,1]$ such that the following holds true
\begin{align}
(1+2\alpha)^K&=1+2K\alpha +\frac{1}{2}K(K-1)(1+\delta(2\alpha)^{K-2}(2\alpha)^2\nonumber\\
& \le 1+2K\alpha +\frac{1}{2}K^2(1+2\alpha)^{K-2}(2\alpha)^2.
\end{align}
Thus, it is clear that
\begin{align}
\frac{(1+2\alpha)^K-2K\alpha-1}{\alpha}\le 2\alpha K^2(1+2\alpha)^{K-2}\label{eq:leT}.
\end{align} %\label{eq:normpc}
Upon plugging \eqref{eq:leT} into \eqref{eq:normpa}, it follows that
\begin{align}
\|\bm{p}(k,\BTheta(k),K)\| &\le 2\alpha K^2(1+2\alpha)^{K-2}(\|\bar{\bbtheta}(k)\|+\frac{r_{\max}}{2}) \nonumber\\
&\le 2\alpha K^2(1+2\alpha)^{K-2}(\|\bar{\bbtheta}(k)-\bbtheta^*\|+\|\bbtheta^*\|+\frac{r_{\max}}{2}). \label{eq:normpb}
\end{align}

Now, we turn to the third term in \eqref{eq:thetade}
\begin{align}
 &\quad  2\alpha \mathbb{E}\bigg[\left\langle \bar{\bbtheta}(k)-\bbtheta^*,\sum_{j=k}^{k+K-1}\frac{1}{M}\Big[\bm{G}^\top(\bm\BTheta(j),\xi_j)\bm{1}-\bm{G}^\top(\bm\BTheta(k),\xi_j)\bm{1}\Big]\right\rangle\Big|\mathcal{F}(k)\bigg]\nonumber \\
 & =2\alpha \mathbb{E}\left [\left\langle\bar{\bbtheta}(k)-\bbtheta^*, \bm{p}(k,\BTheta(k),K)\right\rangle |\mathcal{F}(k) \right] \nonumber\\
 &\le 2\alpha \mathbb{E}\left[\|\bar{\bbtheta}(k)-\bbtheta^*\| \cdot \|\bm{p}(k,\BTheta(k),K)\| \big|\mathcal{F}(k) \right]\nonumber \\
 &=2\alpha \|\bar{\bbtheta}(k)-\bbtheta^*\|\cdot \mathbb{E}\left[ \|\bm{p}(k,\BTheta(k),K)\| \big|\mathcal{F}(k)\right]\nonumber \\
 &\le 4\alpha^2 K^2(1+2\alpha)^{K-2} \|\bar{\bbtheta}(k)-\bbtheta^*\|\cdot (\|\bar{\bbtheta}(k)-\bbtheta^*\|+\|\bbtheta^*\|+\frac{r_{\max}}{2})\nonumber \\
 &\le 4\alpha^2 K^2(1+2\alpha)^{K-2}\Big (2\|\bar{\bbtheta}(k)-\bbtheta^*\|^2+\frac{1}{4}\|\bbtheta^*\|^2 +\frac{r_{\max}}{8}\Big).\label{eq:term3}
\end{align} 
where the second inequality is obtained by plugging in \eqref{eq:normpb}, and the last one follows from the inequality $a(a+b)\le 2a^2+(1/4)b^2$.  

\item[\textbf{3)}] \textbf{Bounding the fourth term.} It follows that
\begin{align}
 &\quad2\alpha \mathbb{E}\bigg[ \left\langle \bar{\bbtheta}(k)-\bbtheta^*,\sum_{j=k}^{k+K-1}\frac{1}{M}\bm{G}^\top(\bm\BTheta(k),\xi_j)\bm{1}-K\bar{\bm{g}}(\bar{\bbtheta}(k)) \right\rangle\Big|\mathcal{F}(k)\bigg]\nonumber \\
 &= 2\alpha \left\langle  \bar{\bbtheta}(k)-\bbtheta^*, \mathbb{E}\bigg[ \sum_{j=k}^{k+K-1}\frac{1}{M}\bm{G}\bm(\BTheta(k),\xi_j))^T\bm{1}-K\bar{\bm{g}}(\bar{\bbtheta}(k))\Big|\mathcal{F}(k) \bigg]\right\rangle\nonumber  \\
 &\le 2\alpha \|\bar{\bbtheta}(k)-\bbtheta^*\| \cdot \Big \|\mathbb{E}\bigg[ \sum_{j=k}^{k+K-1}\frac{1}{M}\bm{G}\bm(\BTheta(k),\xi_j))^T\bm{1}-K\bar{\bm{g}}(\bar{\bbtheta}(k))\Big|\mathcal{F}(k) \bigg] \Big\|\nonumber \\
&\le 2\alpha K \sigma(K)\|\bar{\bbtheta}(k)-\bbtheta^*\|(\|\bar{\bbtheta}(k)-\bbtheta^*\|+1) \nonumber\\
&\le 2\alpha K \sigma(K)\Big (2\|\bar{\bbtheta}(k)-\bbtheta^*\|^2+\frac{1}{4} \Big).\label{eq:term4}
\end{align}

\item[\textbf{4)}] \textbf{Bounding the last term.} Evidently, we have that
\begin{align}
&\quad\Big \|\frac{1}{M}\sum_{j=k}^{k+K-1} \Big[\bm{G}^\top(\bm\BTheta(j),\xi_j)\bm{1}-\bm{G}^\top(\bm\BTheta(k),\xi_j)\bm{1}+\bm{G}^\top(\bm\BTheta(k),\xi_j)\bm{1} \Big]\Big\|^2 \nonumber\\
&\le 2\,\|\bm{p}(k,\BTheta(k),K)\|^2+2\,\Big\|\sum_{j=k}^{k+K-1}\frac{1}{M}\bm{G}^\top(\bm\BTheta(k),\xi_j)\bm{1} \Big\|^2 \nonumber\\
&\le 2\,\|\bm{p}(k,\BTheta(k),K)\|^2+2\,\Big\|\sum_{j=k}^{k+K-1}\bm{H}(j)\bar{\bbtheta}(k)+\frac{1}{M}\bm{r}^{\top}(j)\bm{1}\bm{\phi}(j) \Big\|^2 \nonumber\\
& \le 16\alpha^2 K^4(1+2\alpha)^{2K-4}\|\bar{\bbtheta}(k)\|^2+16K \|\bar{\bbtheta}(k)\|^2 +\big[\alpha^2 K^4(1+2\alpha)^{2K-4}+4K\big]r^2_{\max}\nonumber\\
& \le \Big[32\alpha^2 K^6(1+2\alpha)^{2K-4}+32 K\Big] \big(\|\bar{\bbtheta}(k)-\bbtheta^*\|^2+\|\bbtheta^*\|^2\big)+\big[\alpha^2 K^4(1+2\alpha)^{2K-4}+4K\big]r^2_{\max}\label{eq:term5}
\end{align}
where the first and the last inequality is the result of $\|\sum_{i=1}^n \bm{x}_i\|^2\le n\sum_{i=1}^n\|\bm{x}_i\|^2$ for any $\bm{x}$ and $n$; and the second is obtained by plugging in \eqref{eq:normpb}. Hence, upon taking expectation of both sides of \eqref{eq:term5} conditioning on $\mathcal{F}(k)$,  
we arrive at
\begin{align}
&\quad \alpha^2\mathbb{E}\bigg[\bigg\|\frac{1}{M}\sum_{j=k}^{k+K-1}\Big[\bm{G}^\top(\bm\BTheta(j),\xi_j)\bm{1}-\bm{G}^\top(\bm\BTheta(k),\xi_j)\bm{1}+\bm{G}^\top(\bm\BTheta(k),\xi_j)\bm{1}\Big]\bigg\|^2\Big|\mathcal{F}(k)\bigg] \nonumber\\
&\le \big[32\alpha^4 K^6(1+2\alpha)^{2K-4}+32K\alpha^2\big] \big(\|\bar{\bbtheta}(k)-\bbtheta^*\|^2+\|\bbtheta^*\|^2\big)+\alpha^2 \big[\alpha^2 K^4(1+2\alpha)^{2K-4}+4K\big]r^2_{\max}. \label{eq:4thtermbounds}
\end{align}\vspace{.1cm}
\end{itemize}
We have successfully bounded each of the four terms in \eqref{eq:thetade}.  
Putting now together the bounds in \eqref{eq:term2}, \eqref{eq:term3}, \eqref{eq:term4}, and \eqref{eq:4thtermbounds} into \eqref{eq:thetade}, we finally arrive at
\begin{equation}
\mathbb{E}\Big[\big\|\bar{\bbtheta}(k+K)-\bbtheta^*\big\|^2\big|\mathcal{F}(k)\Big]\le \big[1+2\alpha T \lambda_{\rm max}^{\bar{\bm{H}}}+\alpha\Gamma_1(\alpha,K)\big]\big\|\bar{\bbtheta}(k)-\bbtheta^*\big\|^2+\alpha\Gamma_2(\alpha,K)
\end{equation}
where 
\begin{align}
\Gamma_1(\alpha,K)&=32\alpha^3 K^4(1+2\alpha)^{2K-4}+32K\alpha+8\alpha K^2(1+2\alpha)^{K-2}+4 K \sigma(K)\\
\Gamma_2(\alpha,K)&=\big[32\alpha^3 K^4(1+2\alpha)^{2K-4}+32K\alpha+\alpha K^2(1+2\alpha)^{K-2}\big]\|\bbtheta^*\|^2 \nonumber \\
&\quad +\big[4\alpha^3 K^4(1+2\alpha)^{2K-4}+\frac{1}{2}\alpha K^2(1+2\alpha)^{K-2}+4\alpha K\big]r^2_{\max}+\frac{1}{2} K \sigma(K) 
 \end{align}

%\begin{comment}
%It can be validated that there exists at least a pair $(\alpha_{\rm max}, K_{\mathcal{G}})$, such that $0<1+2\alpha T \lambda_{\rm max}^{\bar{\bm{H}}}+\Gamma_1(\alpha,K)<1$ holds for any $\alpha\in(0,\alpha_{\rm max})$ and $T= K_{\mathcal{G}}$. A simple choice of $(\alpha_{\rm max}, K_{\mathcal{G}})$ is $K_{\mathcal{G}}=\arg_T ~ \sigma(K)=-\frac{1}{4}\lambda_{\rm max}^{\bar{\bm{H}}}$ and $\alpha_{\rm max}=\arg_{\alpha}~ (1+\gamma)^4\alpha^3 K_{\mathcal{G}}^4(1+(1+\gamma)\alpha)^{2K_{\mathcal{G}}-4}+4(1+\gamma)^2\alpha+2(1+\gamma)^2\alpha K_{\mathcal{G}}^2(1+(1+\gamma)\alpha)^{K_{\mathcal{G}}-2} +\alpha T \lambda_{\rm max}^{\bar{\bm{H}}}=\frac{1}{2}\alpha T \lambda_{\rm max}^{\bar{\bm{H}}}$. 
%\end{comment}

From the definition of our multi-step Lyapunov function, we obtain that
\begin{align}
\mathbb{E}\!\left[\mathbb{V}(k+1)-\mathbb{V}(k)\big|\mathcal{F}(k)\right] &=\mathbb{E}
\Big[\big\|\bar{\bbtheta}(k+K)-\bbtheta^*\big\|^2\big|\mathcal{F}(k)\Big]
-\big\|\bar{\bbtheta}(k)-\bbtheta^*\big\|^2 \nonumber\\
&\le \alpha [2K \lambda_{\rm max}^{\bar{\bm{H}}}+\Gamma_1(\alpha,K)]
\big\|\bar{\bbtheta}(k)-\bbtheta^*\big\|^2+\alpha\Gamma_2(\alpha,K)\nonumber\\
&\le \alpha [2K_{\mathcal{G}} \lambda_{\rm max}^{\bar{\bm{H}}}+\Gamma_1(\alpha_{\max},K_{\mathcal{G}})]
\big\|\bar{\bbtheta}(k)-\bbtheta^*\big\|^2+\alpha\Gamma_2(\alpha_{\max},K_{\mathcal{G}})
 \label{eq:Lystepdif}
\end{align}
where the last inequality is due to the fact that functions $\Gamma_1(\alpha,K_{\mathcal{G}}) $ and $\Gamma_2(\alpha,K_{\mathcal{G}})$ are monotonically increasing in $\alpha$. This concludes the proof. %where $[2\alpha \lambda_{\rm max}^{\bar{\bm{H}}}+\Gamma_1(\alpha,K)]<0$.
\end{proof}

\section{Proof of Lemma~\ref{lem:Lyadiff}}\label{append:vandp2}
\begin{proof} It is straightforward to check that
\begin{align}
%     \mathbb{E}\bigg[
     \big\|\bar{\bbtheta}(k+i)-\bbtheta^*\big\|^2  
%     \big]
     &=\Big\|\bar{\bbtheta}(k+i-1)-\bbtheta^*+\frac{\alpha}{M}\bm{G}^{\top}(\bm\BTheta(k+i-1),\xi_{k+i-1})\bm{1}\nonumber\\
     &\quad -\frac{\alpha}{M}\bm{G}^{\top}(\bm{1}(\bm{\bbtheta}^*)^{\top},\xi_{k+i-1})\bm{1}+\frac{\alpha}{M}\bm{G}^{\top}(\bm{1}(\bm{\bbtheta}^*)^{\top},\xi_{k+i-1})\bm{1}\Big\|^2  \nonumber\\
     &\le \|\bar{\bbtheta}(k+i-1)-\bbtheta^*\|^2]+3\alpha^2 \|\bm{H}(k)(\bar{\bbtheta}(k+i-1)-\bbtheta^*)\|^2 \nonumber\\
     &\quad +3\alpha^2 \Big\|\bm{H}(k)\bbtheta^*+\frac{1}{M}\bbphi(s(k))\bm{r}^\top(k)\bm{1}\Big\|^2  \nonumber\\
     & \le (3+12\alpha^2) \|\bar{\bbtheta}(k+i-1)-\bbtheta^*\|^2+6\alpha^2\big[4\|\bbtheta^*\|^2+r_{\max}^2\big] \nonumber\\
     &\le (3+12\alpha^2)^i \|\bar{\bbtheta}(k)-\bbtheta^*\|^2+6\alpha^2\big[4\|\bbtheta^*\|^2+r_{\max}^2\big]\sum_{j=0}^{i-1}(3+12\alpha^2)^j.
     \end{align}
%{\color{black} where $(1+\gamma)$ is the largest absolute value of the eigenvalues of $\bm{H}(k)$ for all $k$.}

As as result, $\mathbb{V}(k)$ can be bounded as 
\begin{equation}
\begin{aligned}
\mathbb{V}(k) &=\sum_{i=0}^{K_{\mathcal{G}}-1}\|\bar{\bbtheta}(k+i)-
\bbtheta^*\|^2 \\
& \le \sum_{i=0}^{K_{\mathcal{G}}-1} (3+12\alpha^2)^i \|\bar{\bbtheta}(k)-\bbtheta^*\|^2+6\alpha^2(4\|\bbtheta^*\|^2+r_{\max})\sum_{i=1}^{K_{\mathcal{G}}-1}\sum_{j=0}^{i-1}(3+12\alpha^2)^j \\
& = \frac{(3+12\alpha^2)^{K_{\mathcal{G}}}-1}{2+12\alpha^2 } \|\bar{\bbtheta}(k)-\bbtheta^*\|^2\\
&\quad +\alpha^2 \frac{6(3+12\alpha^2) \big [(3+12\alpha^2)^{K_{\mathcal{G}}-1}-1 \big ]-6K_{\mathcal{G}}+6}{(2+12\alpha^2 )^2} \big[4 \|\bbtheta^*\|^2+r^2_{\max}\big]
\end{aligned}
\end{equation}

With $c_5:=\frac{(3+12\alpha_{\rm max}^2)^{K_{\mathcal{G}}}-1}{2+3\alpha_{\rm max}^2 }$ and $c_6:=\frac{6(3+\! 12\alpha_{\rm max}^2)\big [ (3+\! 12\alpha_{\max}^2)^{K_{\mathcal{G}}-1}\!-1\big ]\!-6K_{\mathcal{G}}+6}{2+12\alpha_{\rm max}^2} 
 (4 \|\bbtheta^*\|^2+r^2_{\max}\big)$, we conclude that 
\begin{equation}
\mathbb{V}(k)\le c_5 \|\bar{\bbtheta}(k)-\bbtheta^*\|^2+\alpha^2 c_6.
\end{equation}
\end{proof}

\section{Proof of Theorem~\ref{thm:avecon}}\label{append:maravecon}
\begin{proof}
The convergence of $\mathbb{E}\!\left[\big\|\bar{\bbtheta}(k)-\bbtheta^*\big\|^2\right]$ is separately addressed in two phases:\\
1) The time instant $k< k_{\alpha}$, with $k_{\alpha}= \max\{k|\rho^{k}\ge  \alpha\}$, namely, it holds that $\alpha \sigma(K)\le \sigma_k(K)\le \sigma(K)$ for any $k< k_{\alpha}$;\\
2) The time instant $k\ge  k_{\alpha}$, i.e., it holds that $\sigma_{k}(K)\le  \alpha \sigma(K)$ for any $k\ge  k_{\alpha}$.

\textbf{Convergence of the first phase}

From Lemma~\ref{lem:Lyadiff}, we have
\begin{equation}\label{eq:Ld}
-\|\bar{\bbtheta}(k)-\bbtheta^*\|^2 \le -\frac{1}{c_5}\mathbb{V}(k)+\frac{\alpha^2c_6}{c_5}.
\end{equation}
Substituting \eqref{eq:Ld} into \eqref{eq:Lystepdif}, and rearanging the terms give the recursion of Lyapunov function as follows
\begin{align}
\mathbb{E}\!\left[\mathbb{V}(k+1)\big|\mathcal{F}(k)\right]&\le \Big\{1+\frac{1}{c_5} \big[2\alpha K_{\mathcal{G}} \lambda_{\rm max}^{\bar{\bm{H}}}+\alpha \Gamma_1(\alpha_{\max},K_{\mathcal{G}})\big]\Big\} \mathbb{E}\!\left[\mathbb{V}(k)\big|\mathcal{F}(k)\right]\nonumber\\
&\quad +\alpha\Big\{\Gamma_2(\alpha,K_{\mathcal{G}})-\frac{\alpha^2 c_6}{c_5} \big[2 K_{\mathcal{G}} \lambda_{\rm max}^{\bar{\bm{H}}}+\Gamma_1(\alpha_{\max},K_{\mathcal{G}})\big]\Big\}\nonumber \\
&\le c_7 \mathbb{E}\!\left[\mathbb{V}(k)\big|\mathcal{F}(k)\right] +\alpha c_8 \label{eq:Vrecur}
\end{align}
where $c_7:=1+\frac{1}{2c_5} \alpha_{\max} K_{\mathcal{G}} \lambda_{\rm max}^{\bar{\bm{H}}}\in(0,1)$; constant $c_8:=\Gamma_2(\alpha_{\rm max},K_{\mathcal{G}})-\frac{\alpha_{\rm max}^2c_6}{c_5} K_{\mathcal{G}} \lambda_{\rm max}^{\bar{\bm{H}}}>0$, and the last inequality holds true because of \eqref{eq:stepbound}.

Deducing from \eqref{eq:Vrecur}, we obtain that
\begin{align}
\mathbb{E}[\mathbb{V}(k)] &\le c_7^k \mathbb{V}(0)+\alpha c_8\frac{1-c_7^k}{1-c_7} \nonumber\\
&= c_5c_7^k \|\bar{\bbtheta}(0)-\bbtheta^*\|^2+\alpha^2c_6c_7^k+\alpha c_8\frac{1-c_7^k}{1-c_7} \nonumber\\
&\le c_5c_7^k \|\bar{\bbtheta}(0)-\bbtheta^*\|^2+\alpha^2c_6+\frac{\alpha c_8}{1-c_7}\\
&= c_5c_7^k \|\bar{\bbtheta}(0)-\bbtheta^*\|^2+\alpha^2c_6-\frac{ 2c_5c_8}{K_{\mathcal{G}} \lambda_{\rm max}^{\bar{\bm{H}}}}
\end{align}

Recalling the definition of Lyapunov function, it is obvious that 
\begin{equation}
\mathbb{E}\big[\|\bar{\bbtheta}(k)-\bbtheta^*\|^2\big]\le \mathbb{E}[\mathbb{V}(k)]\le  c_5c_7^k \|\bar{\bbtheta}(0)-\bbtheta^*\|^2+\alpha^2 c_6 -\frac{ 2c_5c_8}{K_{\mathcal{G}} \lambda_{\rm max}^{\bar{\bm{H}}}}
\end{equation}
which finishes the proof of the first phase.

\textbf{Convergence of the second phase}\\
Without repeating similar derivation, we directly have that the following holds for $\sigma_k(K)\le \alpha \sigma(K)$:
\begin{align}
&\Gamma_1(\alpha,K):=32\alpha^3 K^4(1+2\alpha)^{2K-4} +32K\alpha +8\alpha K^2(1+2\alpha)^{K-2}+4 K\alpha \sigma(K)\label{eq:gamma1} \\
%\end{align} 
%and
%\begin{align}
&\Gamma_2 (\alpha,K):=\big[32\alpha^3 K^4 (1+2\alpha)^{2K-4}+32K\alpha  
+\alpha K^2(1+2\alpha)^{K-2}\big]\|\bbtheta^*\|^2  \nonumber \\
&\qquad\qquad \qquad +\big[4\alpha^3 K^4(1+2\alpha)^{2K-4}+\frac{1}{2}\alpha K^2(1+2\alpha)^{K-2}+4\alpha K\big]r^2_{\max}+\frac{1}{2} K\alpha \sigma(K).
\end{align}

Subsequently, we have the following recursion of $\mathbb{V}(k)$ that is similar to but slightly different from \eqref{eq:Vrecur}.
\begin{align}
\mathbb{E}\!\left[\mathbb{V}(k+1)\big|\mathcal{F}(k)\right]\le c_7 \mathbb{E}\!\left[\mathbb{V}(k)\big|\mathcal{F}(k)\right] +\alpha^2 c'_8, \quad \forall k\ge k_{\alpha} \label{eq:VVrecur}
\end{align}
where $c'_8:=\big[16\alpha_{\max}^2 K_{\mathcal{G}}^6 (1+2\alpha_{\max})^{2K_{\mathcal{G}}-4}+32K_{\mathcal{G}}  
+2 K_{\mathcal{G}}^3(1+2\alpha_{\max})^{K_{\mathcal{G}}-2}\big]\|\bbtheta^*\|^2+4 K_{\mathcal{G}}r^2_{\max}-\frac{1}{8} K_{\mathcal{G}} \lambda_{\rm max}^{\bar{\bm{H}}} -\frac{\alpha_{\rm max}c_6}{c_5} K_{\mathcal{G}} \lambda_{\rm max}^{\bar{\bm{H}}}$. It is easy to check that $c_8'\ge c_8 $ due to the fact that $\alpha_{\max}<1$ in our case.

Repeatedly applying the above recursion from $k=k_{\alpha}$ to any $k>k_{\alpha}$ yields
\begin{align}
\mathbb{E}[\mathbb{V}(k)] &\le c_7^{k-k_{\alpha}} \mathbb{E}\left[\mathbb{V}(k_{\alpha}) \right]+\alpha^2 c'_8\frac{1-c_7^{k-k_{\alpha}}}{1-c_7} \nonumber\\
&\le  c_7^{k-k_{\alpha}} \Big(c_5c_7^{k_\alpha} \|\bar{\bbtheta}(0)-\bbtheta^*\|^2+\alpha^2c_6-\frac{ 2c_5c_8}{K_{\mathcal{G}} \lambda_{\rm max}^{\bar{\bm{H}}}}\Big)- \alpha\frac{ 2c_5c_8'}{K_{\mathcal{G}} \lambda_{\rm max}^{\bar{\bm{H}}}}\nonumber\\
&\le c_5c_7^{k} \|\bar{\bbtheta}(0)-\bbtheta^*\|^2+c_7^{k-k_{\alpha}}\alpha^2 c_6-(c_7^{k-k_{\alpha}}+\alpha)\frac{ 2c_5c_8'}{K_{\mathcal{G}} \lambda_{\rm max}^{\bar{\bm{H}}}}
\label{eq:lasti}
\end{align}
where we have used $c_8\le c_8'$ for simplicity.

Again, using the definition of the Lyapunov function and \eqref{eq:lasti}, it follows that
\begin{equation}
\mathbb{E}\big[\|\bar{\bbtheta}(k)-\bbtheta^*\|^2\big]\le c_5c_7^{k} \|\bar{\bbtheta}(0)-\bbtheta^*\|^2+c_7^{k-k_{\alpha}}\alpha^2 c_6-(c_7^{k-k_{\alpha}}+\alpha)\frac{ 2c_5c_8'}{K_{\mathcal{G}} \lambda_{\rm max}^{\bar{\bm{H}}}},\quad  \forall k\ge k_{\alpha}
\end{equation}

Combining the results in the above two phases, we conclude that the following bound holds for any $k\in\mathbb{N}^+$
\begin{equation}
\mathbb{E}\big[\|\bar{\bbtheta}(k)-\bbtheta^*\|^2\big]\le c_5c_7^{k} \|\bar{\bbtheta}(0)-\bbtheta^*\|^2-\frac{ 2c_5c_8'}{K_{\mathcal{G}} \lambda_{\rm max}^{\bar{\bm{H}}}}\alpha  + \min\{1, \,c_7^{k-k_{\alpha}}\} \times \left( \alpha^2 c_6-\frac{ 2c_5c_8'}{K_{\mathcal{G}} \lambda_{\rm max}^{\bar{\bm{H}}}}\right).
\end{equation}

\end{proof}

\end{document}